\documentclass[journal]{IEEEtran}
\DeclareUnicodeCharacter{2061}{}
\usepackage{amsmath,amsfonts}
\usepackage{algorithmic}
\usepackage{algorithm}
\usepackage{array}
\usepackage[caption=false,font=normalsize,labelfont=sf,textfont=sf]{subfig}
\usepackage{textcomp}
\usepackage{stfloats}
\usepackage{url}
\usepackage{verbatim}
\usepackage{graphicx}
\usepackage{cite}
\usepackage{algorithm}
\usepackage{algorithmic}

\usepackage{amsmath}
\usepackage{amssymb}
\usepackage{mathtools}
\usepackage{amsthm}

\usepackage{colortbl} 
\usepackage{xcolor}   
\usepackage{xcolor}         
\usepackage{colortbl}

\usepackage{pdfpages}

\newtheorem{theorem}{Theorem}

\newtheorem{lemma}{Lemma}
\newtheorem{corollary}{Corollary}
\theoremstyle{definition}
\newtheorem{definition}{Definition}
\newtheorem{assumption}{Assumption}
\theoremstyle{remark}

\usepackage{lipsum}   
\usepackage{multirow}

\usepackage{booktabs} 
\usepackage{utfsym}
\usepackage{comment}

\begin{document}

\title{Enhanced Byzantine-Robust Federated Learning Via Truncated-Quadratic Loss for Heterogeneous Data}

\author{Zhi-Yong Wang, Hao Nan Sheng, Werner Stefan,~\IEEEmembership{Fellow,~IEEE}, Hing Cheung So,~\IEEEmembership{Fellow,~IEEE}, Linqi Song and Weitao Xu 
		
\thanks{Z.-Y. Wang is with State Key Laboratory of Ocean Sensing, Ocean College, Zhejiang University, Hangzhou, China.


H. N. Sheng and H. C. So are with the Department of Electrical Engineering, City University of Hong Kong, Hong Kong, China. 

Werner Stefan is with the Department of Information and Communications Engineering, Aalto University, 02150 Espoo, Finland.

Linqi Song and Weitao Xu are with the Department of Computer Science, City University of Hong Kong, Hong Kong, China.
(E-mail: zhiyong.wang@aalto.fi, hnsheng2-c@my.cityu.edu.hk, 
stefan.werner@aalto.fi, hcso@ee.cityu.edu.hk, linqi.song@cityu.edu.hk, weitaoxu@cityu.edu.hk). }}



\maketitle

\begin{abstract}
Federated learning distributes data among $n$ clients, making it vulnerable to malicious attacks and data heterogeneity, which together pose challenges for robust learning.
To tackle this issue, centered clipping and Huber aggregators have been exploited for Byzantine robustness.
In this paper, we first demonstrate their equivalence via convex conjugate theory, and show that they can yield biased solutions in the presence of outliers, leading to failure under high data heterogeneity and a substantial fraction of outliers.
Next, we propose a new robust aggregation rule that utilizes the truncated-quadratic (TQ) loss, effectively mitigating the biases of existing methods, such as centered clipping and Huber aggregators. We show that our aggregator achieves order-optimal Byzantine-robust learning under nonconvex loss functions and heterogeneous data, ultimately enhancing the reliability of federated learning systems. 
Additionally, we provide a robust deviation estimation strategy for TQ, demonstrating its effectiveness.
Furthermore, we show that TQ maintains robustness even when only an estimate of the number of Byzantine clients is available. 
Finally, experimental results on MNIST, Fashion-MNIST, and CIFAR-10, indicate that our aggregator provides better robustness performance than the competing techniques. 
\end{abstract}

\section{Introduction}
\IEEEPARstart{F}{ederated} learning (FL) has emerged as a popular paradigm to train large-scale deep learning models with data distributed among $n$ clients such as autonomous vehicles, wearable devices, smart robots and mobile phones, thereby aiming to enhance privacy~\cite{mcmahan2017communication, chen2023energy, yang2024byzantine, huang2024pprp, warnat2021swarm}. 
Each client trains a local model on its own data and sends gradient (or model-update) information to a server, which aggregates these updates to refine a global model.
Although FL has achieved success in numerous fields such as mobile devices \cite{he2025ppbr}, recommendation systems \cite{huynh2025certified} and large language model fine-tuning \cite{wu2024fedbiot}, it is vulnerable to Byzantine clients, which can send arbitrary updates to the server and cause the learning process to diverge~\cite{farhadkhani2022byzantine, blanchard2017machine,zhang2025rethinking}; such behavior is commonly referred to as a Byzantine (malicious) attack \cite{so2020byzantine}.

To resist malicious attacks, Byzantine-resilient FL methods have been studied, and numerous robust aggregation rules have been proposed \cite{guerraoui2018hidden,huang2024self}.
For example, Krum~\cite{blanchard2017machine}, coordinate-wise median (CM)~\cite{yin2018byzantine} and robust federated aggregation (RFA)~\cite{pillutla2022robust} have been developed for robustness against attacks.
However, these rules typically assume data homogeneity, i.e., the data distribution across local datasets are identical, and they become suboptimal under heterogeneous data.
Centered clipping (CC)~\cite{karimireddy2021learning} has been proposed to handle heterogeneous data, but its true breakdown point is unknown.
Here, the term ``breakdown point" refers to the smallest proportion of outliers that can make a robust rule fail. Without additional knowledge about the data, its maximal attainable value is $0.5$ \cite{zoubir2018robust}.
Besides, it has been recently shown that the Huber loss can handle heterogeneous data~\cite{zhao2024huber}, but it requires determining a hyperparameter.

Although CC and Huber have been shown to handle heterogeneous data, their relationship has not been formally examined. 
Moreover, \cite{karimireddy2021learning} demonstrates that CC can resist attacks such as inner product manipulation (IPM)~\cite{xie2020fall} and bit-flipping (BF)~\cite{karimireddy2021learning} with magnitudes no larger than those of good clients.  However, our experimental results show that CC and Huber break down when faced with large-magnitude  outliers, especially under strong data heterogeneity or a high fraction of Byzantine clients. 
The failure mode has not been previously analyzed. 
In this paper, we first establish the relationship between CC and Huber, and provide theoretical analysis of the observed failure.
To address these limitations, we propose a new robust aggregator based on the truncated-quadratic (TQ) loss.
Our main contributions are as follows:
\begin{itemize}
\item[(i)] Using convex-conjugate theory, we show that the Huber loss is equivalent to CC for Byzantine-robust FL, and that both rules satisfy $(f, \kappa)$-robustness. 

\item[(ii)] Since Huber employs the $\ell_1$-norm to resist outliers and is not bounded from above, we establish that it produces a bias that grows with data heterogeneity and the Byzantine fraction; the bias compounds across communication rounds. This analysis explains the empirical failures of Huber and CC when handling outliers under high heterogeneity or a substantial fraction of outliers.

\item[(iii)] We introduce the TQ loss for enhanced outlier resistance. 
To optimize the TQ loss minimization problem, convex conjugate theory is adopted to recast it into its equivalent form.
We prove that the TQ satisfies $(f, \kappa)$-robustness with $\delta_{\min}=0.5$, and we provide a robust deviation estimation strategy for the hyperparameter involved in TQ, analyzing its effectiveness.
Besides, we study the impact of estimating the true number of Byzantine clients on performance, and find that consistently estimating about half of the clients as Byzantine clients still allows our aggregator to maintain robustness.

\item[(iv)] We analyze that TQ achieves order-optimal FL under heterogeneous data, and experimental results using MNIST, Fashion-MNIST and CIFAR-10 show that TQ yields a better Byzantine-resilient performance for most cases.

\end{itemize}

The remainder of this paper is organized as follows. Section \ref{Re_Work} reviews related works and Section \ref{Sec:lim_hu} analyzes the limitations of Huber and CC. The developed robust aggregation rule and robust FL algorithm are presented in Sections \ref{Sec:TQ} and \ref{Sec:TQ-FL}, respectively.
Finally, the performance of the proposed scheme is evaluated in
Section \ref{Sec:Ex_res} and Section \ref{Sec:con} gives the conclusion.
Table \ref{Notation-list} summarizes the notations used throughout.

\begin{table}
		\caption{\small {List of notations.}}  
		\begin{center}
			\setlength{\tabcolsep}{2mm}{
				\begin{tabular}{cc}
					\toprule
					\multicolumn{1}{c}{Notation} &\multicolumn{1}{c}{Definition}\\
                    \hline
                    \hline
                    \multicolumn{1}{c}{$x$} &\multicolumn{1}{c}{Scalar}\\
                    \multicolumn{1}{c}{$\pmb x$} &\multicolumn{1}{c}{Vector}\\
                    \multicolumn{1}{c}{$\pmb X$} &\multicolumn{1}{c}{Matrix}\\
                    \multicolumn{1}{c}{$\|\pmb x\|$} &\multicolumn{1}{c}{Euclidean norm}\\
                    \multicolumn{1}{c}{$n$} &\multicolumn{1}{c}{Number of clients}\\
                    \multicolumn{1}{c}{$f$} &\multicolumn{1}{c}{Number of Byzantine clients}\\
                    \multicolumn{1}{c}{$f_{\max}$} &\multicolumn{1}{c}{Maximum number of Byzantine clients}\\
                    \multicolumn{1}{c}{$\delta:=\frac{f}{n}$} &\multicolumn{1}{c}{Fraction of Byzantine clients}\\
                    \multicolumn{1}{c}{$\delta_{\min}$} &\multicolumn{1}{c}{Breakdown point}\\
                    \multicolumn{1}{c}{$\mathcal G$} &\multicolumn{1}{c}{Set of good clients}\\
                    \multicolumn{1}{c}{$\mathcal B$} &\multicolumn{1}{c}{Set of Byzantine clients}\\
                    \multicolumn{1}{c}{$\kappa$} &\multicolumn{1}{c}{Robustness coefficient}\\
                    \multicolumn{1}{c}{$\pmb w$} &\multicolumn{1}{c}{Model parameter}\\
                    \multicolumn{1}{c}{$b$} &\multicolumn{1}{c}{Batch size}\\
                    \multicolumn{1}{c}{$\mu$} &\multicolumn{1}{c}{Momentum coefficient}\\
                    \multicolumn{1}{c}{$\lambda$} &\multicolumn{1}{c}{Learning rate}\\
                    \multicolumn{1}{c}{$\nabla f$} &\multicolumn{1}{c}{Gradient of $f$}\\
                    \bottomrule             
			\end{tabular}}
			\vspace{-2em}
			\label{Notation-list}
		\end{center}
	\end{table}

 \begin{table*}[htp]
\caption{Comparison among different aggregation rules under Byzantine attacks.}
\label{robust-c1}
\begin{center}
\setlength{\tabcolsep}{1.0mm}{
\begin{tabular}{ccccccc}
    \toprule
    \multicolumn{1}{c}{Aggregator} &\multicolumn{1}{c}{Model} &\multicolumn{1}{c}{Non-convex loss} & \multicolumn{1}{c}{Data type}  & \multicolumn{1}{c}{Breakdown point}&\multicolumn{1}{c}{Robust coefficient}\\
    \hline
    \multicolumn{1}{c}{Krum \cite{blanchard2017machine}} & $\arg\min_{\pmb x_i} \min_{\mathcal{N}_j} \sum_{j\in \mathcal{N}_j}\|\pmb x_i-\pmb x_j\|^2$& \usym{1F5F8}  &IID &$-$
    &$6+\frac{6\delta}{1-2\delta}$\\
    \hline
    \multicolumn{1}{c}{TM \cite{yin2018byzantine}} & $[{\rm TM}(\pmb x_1, \cdots, \pmb x_n)]_j:= \frac{1}{n-2f}\sum_{i\in \mathcal{M}_j}[\pmb x_i]_j$  &\usym{1F5F8} &non-IID&$-$&$\frac{6\delta}{1-2\delta}\left(1+\frac{\delta}{1-2\delta}\right)$\\
    \hline
    \multicolumn{1}{c}{CM \cite{yin2018byzantine}}  & $[{\rm CM}(\pmb x_1, \cdots, \pmb x_n)]_j:= {\rm Med}([\pmb x_1]_j,\cdots, [\pmb x_n]_j)$& \usym{1F5F8}  &IID &$0.5$
    &$4\left(1+\frac{\delta}{1-2\delta}\right)^2$\\
    \hline
    \multicolumn{1}{c}{RFA \cite{pillutla2022robust}} & $\arg\min_{\pmb x} \sum_{i=1}^n\|\pmb x-\pmb x_i\|$& \usym{2613}  &IID &$0.5$
    &$4\left(1+\frac{\delta}{1-2\delta}\right)^2$\\
    \hline
    \multicolumn{1}{c}{CC \cite{karimireddy2021learning}/Huber \cite{zhao2024huber}} & (\ref{CC})/(\ref{Hu-formulation}) & \usym{1F5F8}  &non-IID
    &$0.5$
    &$\frac{4\delta}{1-2\sqrt{\delta(1-\delta)}}$\\
    \hline
    \multicolumn{1}{c}{MCA \cite{luan2024robust}} & $\arg \max_{\pmb x} \sum_{i=1}^{n}G_\sigma(\pmb x-\pmb x_i)$& \usym{2613}  &IID
    &$-$
    &$-$\\
    \hline
    \multicolumn{1}{c}{TQ} & (\ref{TQ-formulation})
    & \usym{1F5F8}  &non-IID
    &$0.5$
    &$\frac{4\delta}{1-2\sqrt{\delta(1-\delta)}}$\\
    \bottomrule
    \end{tabular}}
    \end{center}
    where $\mathcal{N}_j$ is the index set after removing the $f+2$ points whose distance from $\pmb x_i$ are bigger than that in $\mathcal{N}_j$, $\mathcal{M}_j$ refers to the index set at the $j$th dimension that discards the $f$ largest and smallest elements, and $G_\sigma(\cdot)$ is the Correntropy with a Gaussian kernel.
    Besides, non-IID refers to that the aggregator can be proved to handle heterogeneous data in theory. 
    Moreover, we are the first to establish that the breakdown point of CC is $0.5$, while \cite{karimireddy2021learning} considers exploring the true breakdown point of CC as future work.
	\vspace{-0.5em}
\end{table*}
    
\section{Related Works}\label{Re_Work}
To combat malicious attacks from Byzantine clients, Byzantine-robust machine learning algorithms have been exploited~\cite{fang2020local,guerraoui2018hidden,so2020byzantine, bao2024boba,11303260,huang2024self}. These methods primarily adopt robust aggregation rules, including RFA, CM and Krum, to limit the negative impact of attacks. 
Approaches based on suspicion~\cite{xie2019zeno,allen2020byzantine} and redundancy~\cite{rajput2019detox,data2020data} have  also been explored. 
These rules typically require data homogeneity (i.i.d. across clients) and can be suboptimal under heterogeneous data.

However, real-world data across different clients are typically heterogeneous~\cite{yi2024near} as data at each client may be collected under different environments or preferences \cite{zuo2025federated}. To facilitate these rules under data heterogeneity, pre-aggregation steps such as Bucketing~\cite{karimireddy2022byzantinerobust} and nearest neighbor mixing (NNM)~\cite{allouah2023fixing} are suggested. Although pre-aggregation can reduce data heterogeneity and improve the performance of RFA, CM and Krum, it increases the computational cost and can reduce the breakdown point of robust rules~\cite{guerraoui2024byzantine}. 
For example, Bucketing randomly divides $n$ updates received from all clients into $\lceil \frac{n}{s} \rceil$ buckets, each containing at most $s$ elements, and then feeds the mean of each bucket into robust rules. Thus, if the breakdown point of the original robust aggregation rule is $0.5$ and $s=2$, the breakdown point of the resulting scheme with bucketing becomes $0.25$ in the worst case where all contaminated updates are distributed among different buckets.
On the other hand, when the fraction of Byzantine clients, denoted as $\delta:=\frac{f}{n}$, is smaller than $\frac{\kappa}{8+9\kappa}$ where $\kappa$ is the robustness coefficient associated with aggregators, NNM can improve the robustness for heterogeneous data~\cite{guerraoui2024robust}. In other words, NNM can enhance the robustness of aggregators under heterogeneity for a small fraction of Byzantine attacks.
Moreover, post-aggregation step has been exploited for heterogeneous data~\cite{dahan2024fault}, but its performance depends on the prior aggregator.

Furthermore, the $\ell_p$-norm has been adopted for robust ML under heterogeneity~\cite{li2019rsa}, though the analysis requires strongly convex objectives.
A clustering scheme is suggested in \cite{yi2024near}, but the breakdown point of the developed rule is not analyzed. 
Another robust method based on maximum correntropy aggregation (MCA) has been proposed~\cite{luan2024robust}, while its analysis relies on a strongly convex loss function, and the suitability of MCA for heterogeneous data has not been theoretically examined. 
A variant of CC has been proposed in \cite{11241044}, but it still suffers from a biased solution.
Table \ref{robust-c1} compares different robust aggregation techniques in terms of definition, objective loss function, data type, breakdown point and robust coefficient.

\section{Limitations of Huber and CC}\label{Sec:lim_hu}
Before analyzing their limitations, we first clarify the relationship between
Huber aggregation and CC.

Given $n$ inputs, i.e., $\pmb x_1, \cdots, \pmb x_n$, CC aggregates them and produces the next iterate $\pmb v_{k+1}$ at communication round $k$ via
\begin{equation}\label{CC}
        \begin{split}
            \pmb v_{k+1} =  \frac{1}{n}\sum_{i=1}^{n} \pmb v_k + \left(\pmb x_i - \pmb v_k \right)\min\left(1, \frac{\tau_k}{\|\pmb x_i - \pmb v_k\|}\right),
        \end{split}
    \end{equation}
where $\tau_{k}>0$ is the clipping radius. Alternatively,  one can obtain an aggregated vector by minimizing the Huber loss~\cite{zhao2024huber}:
\begin{equation}\label{Hu-formulation}
        \arg \min_{\pmb v} \sum_{i=1}^n \phi_{\rm Huber}(\|\pmb x_i - \pmb v\|), 
        \end{equation}
with $\phi_{\rm Huber}(x) = \begin{cases}
				{x^2}/{2},   &|x|\leq \tau  \\
				\tau x-{\tau^2}/{2},   &|x|>\tau
			\end{cases}$, where $\tau$ is similar to $\tau_{k}$, and $\pmb v$ is the aggregated result.

We now formalize the claim that Huber aggregation and CC are equivalent and share the same robustness guarantee. To this end, we first recall the notion of $(f,\kappa)$-robustness.
\begin{definition}[$(f, \kappa)$-robustness\cite{allouah2023fixing}]\label{def-robustness}
    Given $n$ inputs $\pmb x_1,\cdots,\pmb x_n$, with $f< \delta_{\min} n$, an aggregation rule $R(\cdot)$ is $(f, \kappa)$-robust if
        \begin{equation*}
            \|R(\pmb x_1, \cdots, \pmb x_n)-\overline{\pmb x}\|^2\leq \frac{\kappa}{|\mathcal G|}\sum_{i\in \mathcal G}\|\pmb x_i-\overline{\pmb x}\|^2,
        \end{equation*}
        where $\overline{\pmb x} = \frac{1}{|\mathcal G|}\sum_{i\in \mathcal G} \pmb x_i$, $\mathcal G$ denotes the set of good inputs, $\delta_{\min}$ is the breakdown point, and $\kappa>0$ is the robustness coefficient.
	\end{definition}
    
With this definition in place, we can state the main equivalence result in Theorem \ref{equ-CC-Hubrt}.

\begin{theorem}\label{equ-CC-Hubrt}  
    Huber loss minimization is equivalent to CC for Byzantine-robust FL, and both satisfy $(f, \kappa)$-robustness with $\delta_{\min}=0.5$, $\kappa=\frac{4\delta }{1-2\sqrt{\delta(1-\delta)}}$ and $\tau_k^2 = \frac{1-\delta}{\delta}\left(\|\pmb v_k-\overline{\pmb x}\|^2 + \frac{1}{|\mathcal G|}\sum_{i\in \mathcal G}\|\pmb x_i - \overline{\pmb x}\|^2\right)$, where $\delta=\frac{f}{n}$ is the fraction of Byzantine clients.
\end{theorem}
Proof: See Appendix A of the supplementary material.

\begin{assumption}[Gradient dissimilarity]\label{Ass1-G-gd} 
The variance of local gradients among good clients is upper bounded, i.e.,
        \begin{equation*}
            \frac{1}{|\mathcal G|}\sum_{i\in \mathcal G}\|\pmb x_i-\overline{\pmb x}\|^2\leq \max_i \{ \|\pmb x_i-\overline{\pmb x}\|^2 \} \leq G^2,
        \end{equation*}
        where $G>0$ is a constant.
   \end{assumption}
   
Assumption \ref{Ass1-G-gd} provides a bounded data heterogeneity assumption\cite{zuo2025federated,allouah2023fixing}, and we next present two  examples illustrating that Huber and CC produce biased estimates. 

\textbf{Example $1$.} 
Consider that the gradients of good clients are $\pmb x_1=\cdots=\pmb x_{\frac{1-\delta}{2}n} =\nabla G_1= L \pmb w$ and $\pmb x_{\frac{1-\delta}{2}n+1}=\cdots=\pmb x_{(1-\delta)n} =\nabla G_2= L \pmb w - L \pmb q_1$, respectively, where $\frac{1-\delta}{2}n$ is assumed to be an integer for the sake of analysis, and thus $\overline{\pmb x}=\frac{\nabla G_1+\nabla G_2}{2}=L \pmb w - \frac{L \pmb q_1}{2}$ with $\|\pmb q_1\|=\frac{2G}{L}$, while the remaining points $\pmb x_j$ that satisfy $\|\pmb x_j-\overline{\pmb x}\|>G$ for $j=(1-\delta)n+1,\cdots, n$, are considered as outliers.
Therefore, 
$\frac{1}{\mathcal G}\sum_{i\in \mathcal G}\| \pmb x_i - \overline{\pmb x}\|^2=\frac{\|\nabla G_1-\nabla\overline{G}\|^2+ \|\nabla G_2-\nabla\overline{G}\|^2}{2} = \frac{\|L \pmb w- (L \pmb w - \frac{L \pmb q_1}{2})\|^2+ \|L \pmb w - L \pmb q_1 -(L \pmb w - \frac{L \pmb q_1}{2})\|^2}{2}=\frac{L^2\|\pmb q_1\|^2}{4} = G^2$, which satisfies Assumption \ref{Ass1-G-gd}.


 Even when we provide the true mean value $\overline{\pmb x}=\frac{1}{\mathcal G}\sum_{i\in \mathcal G} \pmb x_i$ as the initialization point of Huber and CC with $\tau_0  = \sqrt{\frac{1-\delta}{\delta}\left(\left\|\pmb v_{0}-\overline{\pmb x} \right\|^2 +\frac{1}{|\mathcal G|}\sum_{i\in\mathcal G}\left\|\pmb x_i - \overline{\pmb x}\right\|^2\right)}=\sqrt{\frac{1-\delta}{\delta}G^2}>G $, their first iteration is 
\allowdisplaybreaks[4] 
        \begin{align}
            \pmb v_{1} =& \frac{1}{n}\sum_{i=1}^{n} \pmb v_0 + \left(\pmb x_i - \pmb v_0 \right)\min\left(1, \frac{\tau_0}{\|\pmb x_i - \pmb v_0\|}\right) \nonumber\\ 
            =&(1-\delta)\overline{\pmb x} + \frac{1}{n}\sum_{i\in \mathcal B} \pmb v_0 + \left(\pmb x_i - \pmb v_0 \right)\frac{\tau_0}{\|\pmb x_i - \pmb v_0\|} \nonumber\\
            =&(1-\delta)\overline{\pmb x} + \delta\overline{\pmb x} + \frac{1}{n}\sum_{i\in \mathcal B} \frac{\tau_0 \left(\pmb x_i - \pmb v_0 \right)}{\|\pmb x_i - \pmb v_0\|} \nonumber\\
            =&\overline{\pmb x} + \frac{1}{n}\sum_{i\in \mathcal B} \frac{\tau_0 \left(\pmb x_i - \overline{\pmb x} \right)}{\|\pmb x_i - \overline{\pmb x}\|}, \nonumber
        \end{align}
where $\mathcal B$ is the set of outliers and thus
\begin{equation*}
        \begin{split}
            \|\pmb v_{1}-\overline{\pmb x}\|=& \delta \tau_0 \left\|\frac{1}{|\mathcal B|}\sum_{i\in \mathcal B} \frac{ \pmb x_i - \overline{\pmb x} }{\|\pmb x_i - \overline{\pmb x}\|}\right\|\geq 0.
        \end{split}
    \end{equation*}
That is to say, even if $\pmb v_0=\overline{\pmb x}$, Huber and CC can make the aggregated result deviate the true value at the first iteration.

\begin{figure}[htb]
\footnotesize
  \centering            
  \begin{minipage}[t]{0.48\linewidth} 
      \centering
      \includegraphics[width=\linewidth]{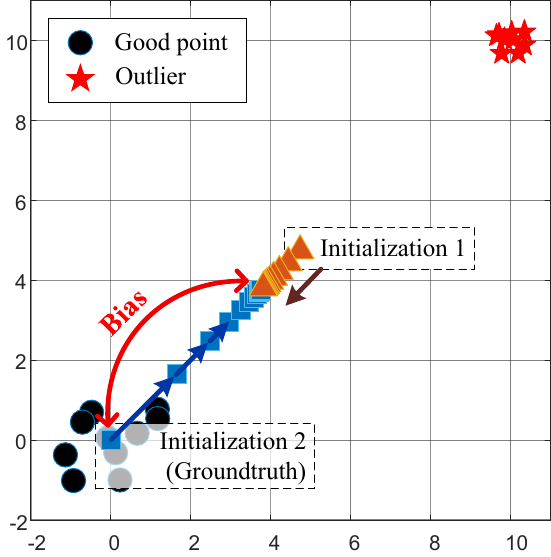} 
      \centerline{(a). Huber loss}
      \label{fig:Huber}
  \end{minipage}
  \hfill 
  \begin{minipage}[t]{0.48\linewidth} 
      \centering
      \includegraphics[width=\linewidth]{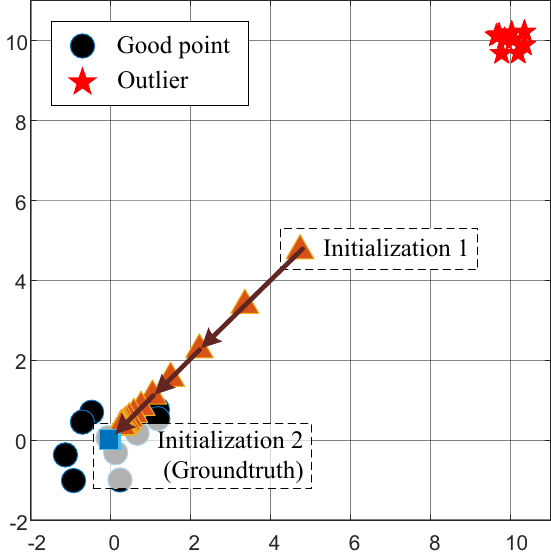} 
      \centerline{(b). TQ loss}
      \label{fig:TQ}
  \end{minipage}
  \vspace{-0.1cm}
  \caption{Iteration process of Huber and TQ loss. Black dots represent good input points, while red pentagrams indicate outliers. Orange triangles indicate the iterative process starting from the average of all points, while blue squares represent the process beginning with the average of good points.}    
  \label{fig SD xi RMSE}         
\end{figure}

\textbf{Example $2$.}
Consider the FL scenario where there are $f=9$ Byzantine clients among $n=19$ clients and the dimension of each vector $\pmb x_i~(i=1,\cdots,19)$ sent by the clients is $2$. That is, $\pmb x_1,\cdots,\pmb x_{19}\in \mathbb{R}^2$.
The good points $\pmb x_i,~i\in \mathcal{G}$ with $|\mathcal{G}|=10$ in black in Fig. \ref{fig SD xi RMSE} are randomly generated with the mean and the variance being $(0,0)$ and $1$, respectively, while the red pentagrams denote the vector sent by the Byzantine clients and the mean and variance of those points are $(10,10)$ and $0.1$, respectively.
Two different initialization points, i.e., the true mean of good points and the mean of all points, are considered, and both Huber and TQ conduct $L=100$ iterations.
Fig. \ref{fig SD xi RMSE} (a) plots the iterates of the Huber aggregator. 
Even when initialized at the true mean of the good points, $(0,0)$, the sequence converges to a biased point. 

{Motivated by Examples 1--2, we theoretically analyze the bias yielded by Huber and CC in Theorem \ref{bias-CC-Huber}. Notably, the resulting bias is strictly positive even under a small fraction of outliers, as long as the data are heterogeneous.
It is demonstrated that when handling outliers, the error between $\overline{\pmb x}$ and the limit point $\pmb v_\star$ increases with the degree of heterogeneity $\frac{1}{\mathcal G}\sum_{i\in \mathcal G}\| \pmb x_i - \overline{\pmb x}\|^2$ and the fraction of Byzantine clients $\delta$.}

\begin{theorem}\label{bias-CC-Huber}  
Suppose there exist $\delta n$ outlier-contaminated inputs among $\pmb x_1,\ldots,\pmb x_n$, and Assumption~\ref{Ass1-G-gd} holds. Let $\mathcal B$ denote the set of outliers with $|\mathcal B|=\delta n$, and define
 $\Delta_m = \left\| \frac{1}{|\mathcal B|}\sum_{i\in \mathcal B} \frac{\pmb x_i - \pmb v_\star}{\|\pmb x_i - \pmb v_\star\|}\right\|$ where $\pmb v_\star:=R(\pmb x_1, \cdots, \pmb x_n)$ is the aggregated result provided by Huber and CC, then
the aggregation error $\|\pmb v_\star- \overline{\pmb x}\|^2=\frac{\delta \Delta_m^2}{1-(1+\Delta_m^2)\delta} \frac{1}{|\mathcal G|}\sum_{i\in \mathcal G}\| \pmb x_i - \overline{\pmb x}\|^2$ if $\delta \leq \frac{2(1+\Delta_m)+\kappa_0 -\sqrt{4\kappa_0 (\Delta_m -\Delta_m^2)+\kappa_0^2}}{2\left( (1+\Delta_m)^2+\kappa_0(1+\Delta_m^2)\right)}$ with $\kappa_0=\frac{G}{\sqrt{\frac{1}{|\mathcal G|}\sum_{i\in \mathcal G}\| \pmb x_i - \overline{\pmb x}\|^2}}$.
\end{theorem}
Proof: See Appendix B of the supplementary material.

\section{Robust Aggregation via TQ}\label{Sec:TQ}
 Section \ref{Sec:lim_hu} shows that Huber yields biased solutions.
The reason is that Huber employs the $\ell_1$-norm to resist outliers, which can  still be impacted by the magnitudes of outliers since Huber loss increases with the magnitudes of outliers.
To address this problem, we adopt TQ for enhanced outlier resistance, defined as 
$$\phi_{\rm TQ}(x) = \begin{cases}
				{x^2}/{2},   &|x|\leq \tau  \\
				{\tau^2}/{2},   &|x|>\tau
			\end{cases}.$$
Similar to Huber, TQ employs the quadratic function for inlier noise, but it clips outliers whose magnitudes exceeds threshold $\tau$.
The aggregation problem for TQ loss is formulated as 
\begin{equation}\label{TQ-formulation}
    \min_{\pmb v} \sum_{i=1}^n \phi_{\rm TQ}(\|\pmb x_i - \pmb v\|).
\end{equation}

Since TQ loss is nonsmooth and nonconvex, (\ref{TQ-formulation}) is converted into its equivalent form using Lemma \ref{TQ-l0}, i.e.,
\begin{equation}\label{OGC}
        \begin{split}
            \pmb v_{k+1} = \pmb v_k + \frac{1}{n}\sum_{i=1}^{n}\left(\pmb x_i - \pmb v_k \right)I_{k,i}(\pmb v_k, \pmb x_i),
        \end{split}
    \end{equation}
where $I_{k,i}(\pmb v_k, \pmb x_i) := 
            \begin{cases} 
                1, & \text{if } \|\pmb x_i - \pmb v_k\| \leq \tau_k \\ 
                0, & \text{otherwise} 
            \end{cases}$, and the proof from (\ref{TQ-formulation}) to (\ref{OGC}) is provided in Appendix C of the supplementary material.

\begin{lemma}[\!\!\!\cite{wang2022robust}]\label{TQ-l0}
        By convex conjugate theory, $\phi_{\rm TQ}(x)$ is equivalent to
        \begin{equation*}\label{TQ-reformulation}
        \phi_{\rm TQ}(x):= \min_y \frac{1}{2}(x-y)^2 + \varphi(y)
        \end{equation*}
        where $\varphi(y)= \begin{cases}
				-\frac{(|y|-\tau)^2}{2} +\frac{\tau^2}{2},   &|y|\leq \tau  \\
				\frac{\tau^2}{2},   &|y|>\tau
			\end{cases}$, 
    and the solution for $y$ is given by the thresholding operator $\mathcal{P}(x) = \begin{cases}
				0,   &|x|\leq \tau  \\
				x,   &|x|>\tau
			\end{cases}$.
    \end{lemma}
    
Algorithm \ref{Algo:OGC} summarizes the steps of the TQ aggregation rule.
Theorem~\ref{OGC_BP} below shows that this rule is $(f, \kappa)$-robust.
Furthermore, replacing Huber with TQ in Example $1$ results in $\|\pmb v_{1}-\overline{\pmb x}\|=0$ and Fig. \ref{fig SD xi RMSE} (b) confirms that as the iterates of the TQ converge toward the true mean value of good points. 

\begin{algorithm}[t]
		\caption{Aggregation rule TQ}
		\label{Algo:OGC}
		\begin{algorithmic}[1]        
		\REQUIRE  Input vectors $\{\pmb x_1, \cdots, \pmb x_n\}$, $\delta$, $k_0$
		\STATE\textbf{Initialize:} $\pmb v_0$ and $k=0$
		\WHILE {$k\leq k_0$}
		\STATE Compute $w_{k,i}=I_{k,i}(\pmb v_k, \pmb x_i)$ 
		\STATE Compute $\pmb v_{k+1} = \pmb v_k + \frac{1}{n}\sum_{i=1}^{n}\left(\pmb x_i - \pmb v_k \right)w_{k,i}$		
		\STATE $k\leftarrow k+1$		
		\ENDWHILE
		\ENSURE $\pmb v_{k_0}$.
		\end{algorithmic}
	\end{algorithm}

\begin{theorem}\label{OGC_BP}
    Given $n$ inputs $\pmb x_1,\cdots,\pmb x_n$ with $|\mathcal G|= (1-\delta)n$ good vectors that are independent and bounded, and define $D_k=\|\pmb v_k-\overline{\pmb x}\|$.
    Starting from any bounded initialization ${\pmb v_0}$, performing TQ for $k$ iterations with $\tau_k^2 = \frac{1-\delta}{\delta}\left(D_k^2 + \frac{1}{|\mathcal G|}\sum_{i\in \mathcal G}\|\pmb x_i - \overline{\pmb x}\|^2\right)$ ensures $D_k\leq \left(\delta+\sqrt{\delta(1-\delta)}\right)^k D_0 + \frac{\sqrt{\delta(1-\delta) \frac{1}{|\mathcal G|}\sum_{i\in \mathcal G}\|\pmb x_i - \overline{\pmb x}\|^2}}{1-\left(\delta+\sqrt{\delta(1-\delta)}\right)}$.
    Moreover, when $\delta < \delta_{\min} = 0.5$, we achieve the following bound after $k= \log_a\frac{\sqrt{\delta(1-\delta)\frac{1}{|\mathcal G|} \sum_{i\in\mathcal G}\|\pmb x_i - \overline{\pmb x}\|^2}}{(1-a)D_0}$ iterations with $a=\delta+\sqrt{\delta(1-\delta)}$:
        \begin{equation*}
            \|\pmb v_k - \overline{\pmb x}\|^2\leq \frac{4\delta }{1-2\sqrt{\delta(1-\delta)}}\frac{1}{|\mathcal G|}\sum_{i\in \mathcal G}\|\pmb x_i-\overline{\pmb x}\|^2.
        \end{equation*}
    \end{theorem}
Proof: See Appendix D of the supplementary material.

The value of the hyperparameter $\tau_k$ is required for the execution of Algorithm~\ref{Algo:OGC}. 
According to Theorem~\ref{OGC_BP}, the calculation of $\tau_k$ requires knowing $\overline{\pmb x}$ and the variance of gradients sent by good clients, which are usually unknown.
To solve this issue, the median and the dispersion in robust statistics are introduced, which have been widely adopted to estimate the mean and the standard deviation of data points that may be contaminated by outliers~\cite{tyler2008robust,zoubir2018robust}. 
The median operator $\rm Med(\cdot)$ is used to estimate the mean, and the median absolute deviation (MAD) is a commonly-used dispersion estimator~\cite{tyler2008robust,zoubir2018robust} to estimate the standard deviation, which is computed as 
    \begin{equation}\label{MAD_V}
        \begin{split}
            {\rm MAD}(\pmb p):={\rm Med}(|\pmb p - {\rm Med}(\pmb p)|),
        \end{split}
    \end{equation}
where $\pmb p:=\{p_1,\cdots,p_n\}$ includes $n$ data points.

Accordingly, for the collected gradients in the server, namely, $\pmb x_1,\cdots, \pmb x_n\in \mathbb{R}^d$, we arrange them into a matrix $\pmb X=[\pmb x_1,\cdots,\pmb x_n]\in \mathbb{R}^{d\times n}$, and estimate $\overline{\pmb x}$ and the variance by 
    \begin{equation}\label{R-est}
        [\hat{\overline{\pmb x}}]_j= {\rm Med}(X_{j,:}),~~\hat{V}=\sum_{j=1}^d \left({\rm MAD}(\pmb X_{j,:})\right)^2,
    \end{equation}
where $[\hat{\overline{\pmb x}}]_j$ and $\pmb X_{j,:}$ denote the $j$th entry of the estimate of $\overline{\pmb x}$ and the $j$th row of $\pmb X$, respectively. 
Thus, the estimate of $\tau_k$ can be obtained.
{Moreover, we analyze that (\ref{MAD_V}) provides a valid dispersion estimate; see Appendix E of the supplementary material. Empirical evidence for the effectiveness of (\ref{MAD_V}) is shown in Fig. \ref{fig-MAD}.}

{Furthermore, it is worth pointing out that Huber and TQ require knowing the number of Byzantine clients, while it is not easy to estimate the value of $f$ for some cases.
Then, it is necessary to analyze that whether inputting $f\leq \hat f \leq f_{\max}$ still makes Huber and TQ satisfy $(f, \kappa)$-robustness. 
Here, $\hat f$ refers to the estimated value of Byzantine workers. We assume $\hat f\geq f$, and if not, the algorithm will consider that there only exist $\hat f$ outliers and may misclassify at least $f-\hat f$ outliers as normal data, leading to divergence.
Theorem \ref{Hubrt-TQ-robust-f} analyzes that Huber and TQ still achieve robustness when inputting $\hat f$ but the corresponding robustness coefficients are bigger than that when inputting the true number of Byzantine client, leading to an increase in the aggregation error.
This is because the discrepancy between $f_{\max}$ and the true value of byzantine clients introduces extra error.
In practice, we can assume $\hat f=f_{\max}$ where $f_{\max}$ refers to the maximum number of Byzantine clients one robust rule can resist and its value relies on the breakdown point of the aggregator.
For instance, if $n=11$ and the breakdown point is $0.5$, then $f_{\max}=5$.
}
{\begin{theorem}\label{Hubrt-TQ-robust-f}
   If, instead of the true number of Byzantine clients, inputting $\hat f$ with $n\ge 2\hat f+1$, then Huber and TQ still satisfy $(f,\kappa)$-robustness; however, their robustness coefficients increase with $\hat f$, resulting in a large aggregation error.
\end{theorem}}
Proof: See Appendix F of the supplementary material.

\section{Robust FL via TQ}\label{Sec:TQ-FL}
\subsection{Proposed Algorithm}

FL considers a total $N$ data points distributed across $n$ clients to train a global model. 
Each client $i$ possesses its own dataset $\mathcal{P}_i$ that includes $n_i$ data points, and its local loss function is given by
\begin{equation*}
    \begin{split}
        \mathcal{F}_i(\pmb w)=\frac{1}{n_i}\sum_{j=1}^{n_i}f(\pmb w, p_j),
    \end{split}
\end{equation*}
where $f(\cdot)$ is the sample loss, $\pmb w\in \mathbb{R}^d$ is the model parameter, and $p_j$ is the $j$th data point in $\mathcal{P}_i$.
To reduce complexity, instead of computing gradient on the whole dataset, each client $i$ randomly samples $b$ data points from its own dataset $\mathcal{P}_i$ to compute local stochastic gradient $\nabla f_i^t=\frac{1}{b}\sum_{j\in \mathcal{P}_i^b}\nabla f(\pmb w_t, p_j)$ where $\mathcal{P}_i^b$ is the set of the sampled $b$ data points.
The computed $\nabla f_i^t$ with $i=1,\cdots, m$ are sent to the server, which aggregates the received gradients and then updates the global model. 
Finally, the server broadcasts the updated global model to each client to start a new round of  updates.
Since up to $f$ Byzantine clients may send arbitrary information to derail training, we need to employ a robust aggregator at the server.

Byzantine-robust FL typically relies on stochastic gradient descent (SGD), which induces variance in the gradient estimates of individual clients.
To address this challenge, several variance reduction schemes have been proposed~\cite{fedin2023byzantine,wu2020federated,zhu2022byzantine}.
This work utilizes the momentum technique, which not only diminishes variance but also helps combat time-coupled attacks~\cite{karimireddy2021learning}. 
Instead of transmitting gradients, clients maintain momentum vectors $\pmb g_i$ and share them with the server; the server aggregates these vectors using the TQ rule to update the global model.
The process can be described as
$$\pmb g_i^t = \mu \pmb g_i^{t-1} + (1-\mu) \nabla f_i^t,$$
$$\pmb w^t = \pmb w^{t-1} - \lambda \cdot {\rm TQ}(\{\pmb g_i^t,\cdots, \pmb g_n^t\}),$$
where $\mu\in[0,1)$ is the momentum coefficient  and $\lambda>0$ is the learning rate.  Algorithm~\ref{Algo:ROGC} provides the detailed steps.

\begin{algorithm}[t]
		\caption{Byzantine-Robust FL}
		\label{Algo:ROGC}
		\begin{algorithmic}[1]
			\REQUIRE  Initial weight parameter $\pmb w^0$, momentum vector $\pmb g_0^i=\pmb 0$, batch size $b$, momentum coefficient $\mu$, learning rate $\lambda$ and total number of strps $T$
			\WHILE {$t= 1,\cdots,T$}
            \STATE Server sends the model parameter $\pmb w^{t-1}$ to all clients
            \FOR{ $i= 1, \cdots, n$ \textbf{parallelly} }
            \IF{$i \in \mathcal{G}$}
                \STATE $\nabla f_i^t=\frac{1}{b}\sum_{j\in \mathcal{P}_i^b}\nabla f(\pmb w_t, p_j)$ 
                \STATE $\pmb g_i^t = \mu \pmb g_i^{t-1} + (1-\mu) \nabla f_i^t$
            \ELSE
                \STATE $\pmb g_i^t $ is any erroneous $d$ dimensional vector
            \ENDIF
            \STATE Send all $\pmb g_i^t$ to the server
            \ENDFOR
			\STATE Server employs TQ aggregator to merge all inputs and then updates parameter by $\pmb w^t = \pmb w^{t-1} - \lambda \cdot {\rm TQ}(\{\pmb g_i^t,\cdots, \pmb g_n^t\})$ 		
			\STATE $t\leftarrow t+1$
			\ENDWHILE
            \ENSURE $\hat{\pmb w}$ randomly sampled from $\{\pmb w^1,\cdots,\pmb w^T\}$.
		\end{algorithmic}
	\end{algorithm}

\subsection{Theoretical Analysis}
As the loss is typically nonconvex, the goal of the server is to obtain a stationary point of $\mathcal F_{\mathcal G}=\frac{1}{|\mathcal G|}\sum_{i\in \mathcal G} \mathcal F_i(\pmb w)$, as formalized in Definition \ref{fe-def}. 
Lemma \ref{Low-bound0} below further demonstrates that the learning error is lower bounded and increases with the fraction of adversaries and the degree of heterogeneity.

\begin{definition}[($f, \varepsilon$)-resilience~\cite{allouah2023fixing}]\label{fe-def}
A Byzantine-robust learning algorithm is ($f, \varepsilon$)-resilient if its model parameter $\hat{\pmb w}$ fulfills the condition $\|\nabla \mathcal F_{\mathcal G}(\hat{\pmb w})\|^2\leq \varepsilon$ in the presence of $f$ adversaries.
\end{definition}

\begin{lemma}[\!\!\!\cite{guerraoui2024robust}]\label{Low-bound0}
Given that Assumptions \ref{Ass1-G-gd}-\ref{Ass3} are satisfied, and $0<\delta<0.5$, a learning algorithm is considered to be ($f, \varepsilon$)-resilient if $\varepsilon\geq\frac{1}{4}\delta G^2$.
\end{lemma}

To analyze the convergence of our algorithm, three common assumptions are imposed, namely, the lower bound of $\mathcal{F}_{\mathcal{G}}$, the Lipschitz continuity of
gradients and the bounded inner gradient variance~\cite{JMLR:v26:24-1307}. 

\begin{assumption}\label{Ass2}   
The loss $\mathcal{F}_{\mathcal{G}}(\pmb w)$ is lower bounded by $\mathcal{F}_{\mathcal{G}}^*$, i.e., $\mathcal{F}_{\mathcal{G}}(\pmb w)\geq\mathcal{F}_{\mathcal{G}}^*$ for any $\pmb w \in \mathbb{R}^d$.
\end{assumption}
\begin{assumption}\label{Ass4} 
The gradient of loss $\mathcal{F}_i(\cdot)$ is $L$-Lipschitz continuous for any $\pmb w_1, \pmb w_2 \in \mathbb{R}^d$, namely, $\|\nabla \mathcal{F}_i(\pmb w_1) - \nabla \mathcal{F}_i(\pmb w_2)\|\leq L\|\pmb w_1 - \pmb w_2\|$.
\end{assumption}

\begin{assumption}\label{Ass3}  
The variance of the stochastic gradient for any good client is upper-bounded by $\mathbb E_{\mathcal{P}_i^b}\|\nabla f_i^t(\pmb w)-\nabla \mathcal F_i(\pmb w)\|^2\leq \sigma^2/b$.
\end{assumption}

Some studies \cite{karimireddy2021learning,allouah2023fixing,guerraoui2024robust} have investigated the convergence of Byzantine-robust FL.
For example, \cite{allouah2023fixing} provides its convergence analysis based on stochastic gradient descent, namely, $b=1$ in Algorithm~\ref{Algo:ROGC}, while \cite{guerraoui2024robust} employs minibatch gradient descent to compute the gradient at each client and analyzes the corresponding convergence under the condition that robust aggregators satisfy Definition \ref{def-robustness}.
In this paper, minibatch gradient descent is adopted; thus, the convergence analysis in \cite{guerraoui2024robust} is suitable for the convergence of Algorithm~\ref{Algo:ROGC} since the developed aggregator TQ satisfies Definition \ref{def-robustness}, and the conclusions are presented in Theorem~\ref{Convergence-OGC}. For the proofs, please refer to Theorem 6.1 in~\cite{guerraoui2024robust}.

\begin{theorem}\label{Convergence-OGC}
    Provided that Assumptions \ref{Ass1-G-gd}-\ref{Ass3} hold, run Algorithm~\ref{Algo:ROGC} for $T>1$ iterations with learning rate $\lambda=\min\left\{\frac{1}{24L}, \frac{1}{8}\sqrt{\frac{\left(\mathcal{F}_{\mathcal{G}}(\pmb w^0)- \mathcal{F}_{\mathcal{G}}^*\right)b|\mathcal{G}|}{\left(3\kappa|\mathcal{G}|+1\right)\sigma^2LT}}\right\} $ and momentum coefficient defined as $\mu=\sqrt{1-24L\lambda}$, then when $T\geq \frac{9\left(\mathcal{F}_{\mathcal{G}}(\pmb w^0)- \mathcal{F}_{\mathcal{G}}^*\right)Lb|\mathcal{G}|}{\left(3\kappa|\mathcal{G}|+1\right)\sigma^2}$, we have
    \begin{equation*}
        \begin{split}
            &\frac{1}{T}\sum_{t=1}^T \mathbb{E}\left[\|\nabla \mathcal{F}_{\mathcal{G}}(\pmb w^{t-1})\|^2\right]\\
            \leq &144\sqrt{\left(3\kappa+\frac{1}{|\mathcal{G}|}\right)\frac{\left(\mathcal{F}_{\mathcal{G}}(\pmb w^0)- \mathcal{F}_{\mathcal{G}}^*\right)L\sigma^2}{Tb}}+36\kappa G^2.
        \end{split}
    \end{equation*}
\end{theorem}

\begin{corollary}\label{cor-t1}
According to Theorem \ref{Convergence-OGC},
the server selects a parameter at random from the set $\{\pmb w^0,\cdots,\pmb w^{T-1}\}$,  then the resultant parameter $\hat{\pmb w}$ in expectation is given by
\begin{equation*}
        \begin{split}
            \mathbb{E}\left[\|\nabla \mathcal{F}_{\mathcal{G}}(\hat{\pmb w})\|^2\right]&=\frac{1}{T}\sum_{t=1}^T \mathbb{E}\left[\|\nabla \mathcal{F}_{\mathcal{G}}(\pmb w^{t-1})\|^2\right]\\
            &\in \mathcal{O}\left(\sqrt{\left(\kappa+\frac{1}{|\mathcal{G}|}\right)\frac{\sigma^2}{Tb}}+\kappa G^2\right).
        \end{split}
    \end{equation*}
\end{corollary}

Corollary~\ref{cor-t1} implies that 
the proposed algorithm achieves ($f, \varepsilon$)-resilience with $\varepsilon\in \mathcal{O}\left(\sqrt{\left(\kappa+\frac{1}{|\mathcal{G}|}\right)\frac{\sigma^2}{Tb}}+\kappa G^2\right)$, and when $T\rightarrow\infty$, the asymptotic error of our algorithm is $\mathcal{O}(\delta  G^2)$ because of $\kappa=\mathcal{O}(\delta)$ for TQ by Theorem \ref{OGC_BP}, which is order-optimal since it aligns with the learning error described in Lemma \ref{Low-bound0}.  
In this context, $order$-$optimality$ signifies that no FL algorithm can attain a learning error below $\delta G^2$ multiplied by a specific factor~\cite{guerraoui2024byzantine}.

\begin{figure}[htbp]    
  \centering            
  {
      \label{fig SD xi RMSE1}\includegraphics[height=0.35\linewidth]{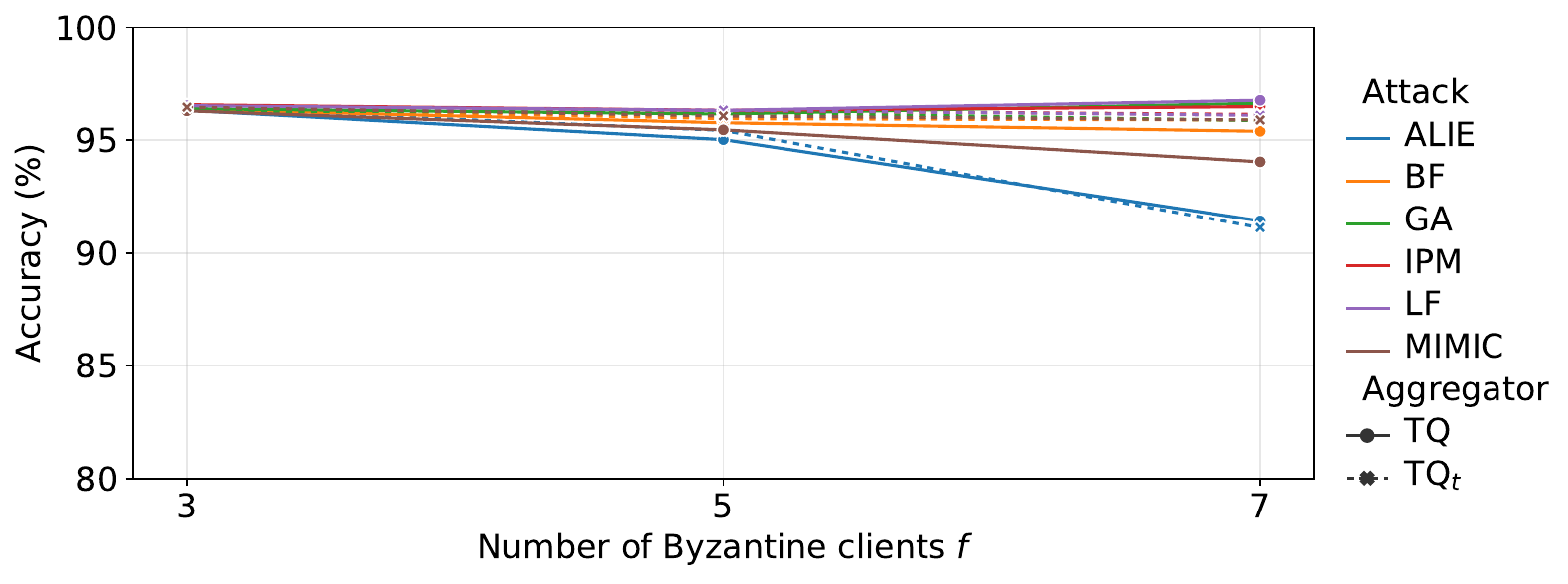}
  } 
  
  {
      \label{fig SD xi RMSE2}\includegraphics[height=0.35\linewidth]{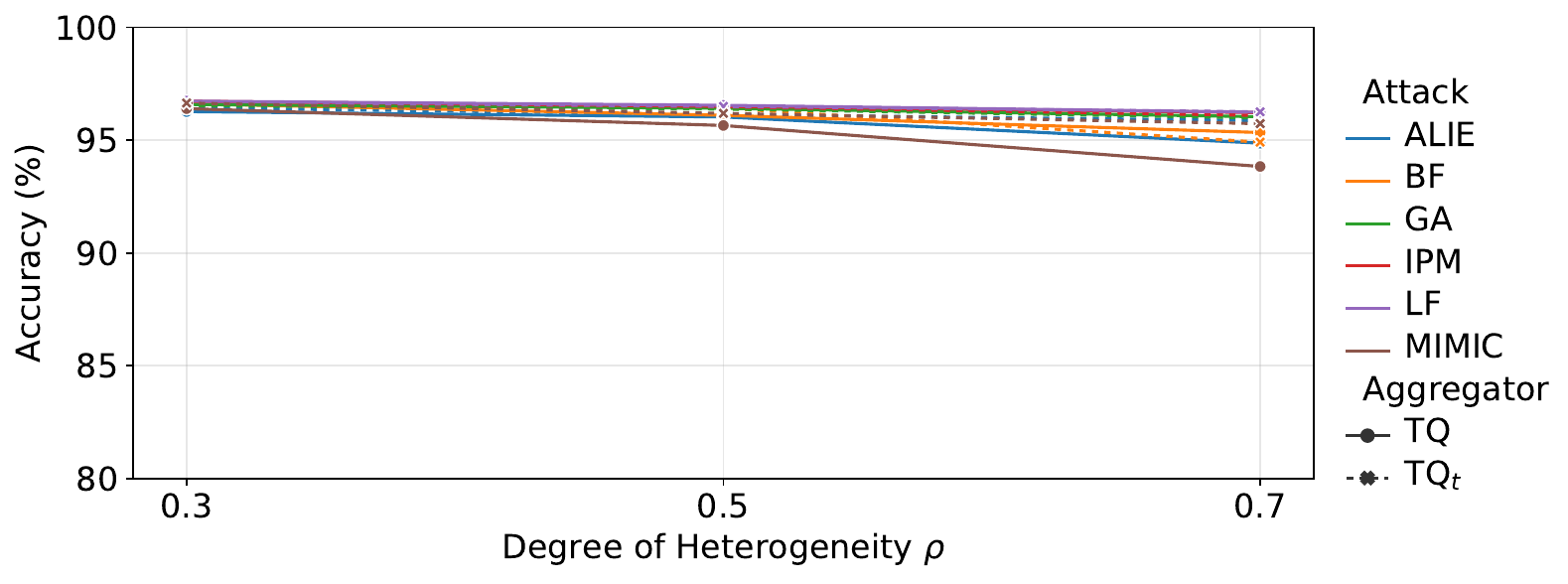}
  }
  \caption{{Effectiveness of robust estimation strategy for hyperparameter $\tau_k$ under different numbers of Byzantine clients and data heterogeneity. Here, TQ computes $\tau_k$ by  (\ref{R-est}) while we provide the true value of  $\tau_k$ for TQ$_t$.}}   
  \label{fig-MAD}            
\end{figure}

\begin{figure*}[h]
\begin{center}
\includegraphics[width=0.95\linewidth]{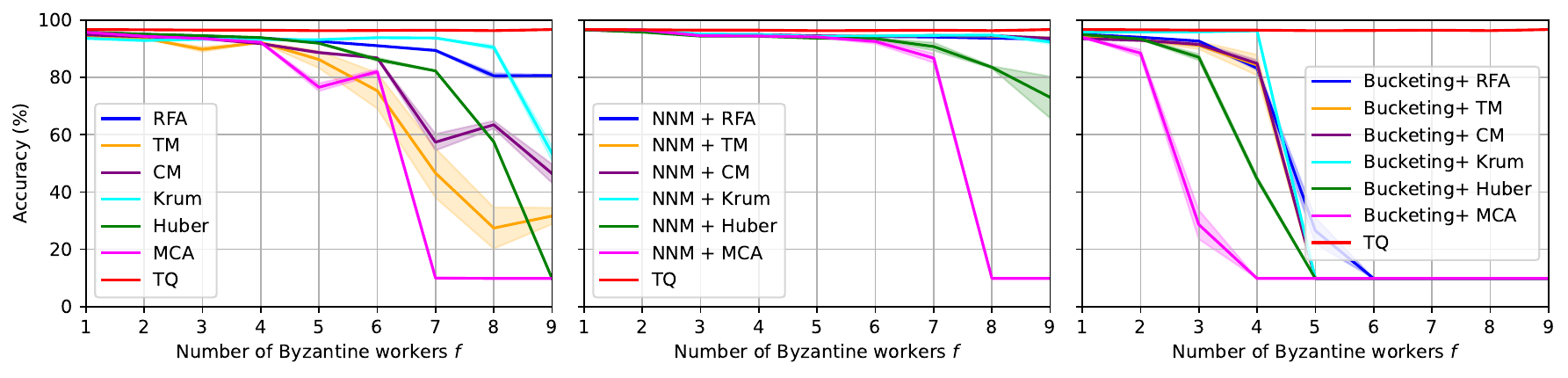}
\end{center}
\caption{Classification accuracy versus number of Byzantine clients for aggregators with or without pre-aggregation under heterogeneity ratio $\rho=0.5$. Left: aggregators only; middle: aggregators with NNM; right: aggregators with Bucketing.}
\label{fig:number-By}
\end{figure*}

\begin{figure*}[h]
\begin{center}
\includegraphics[width=0.99\linewidth]{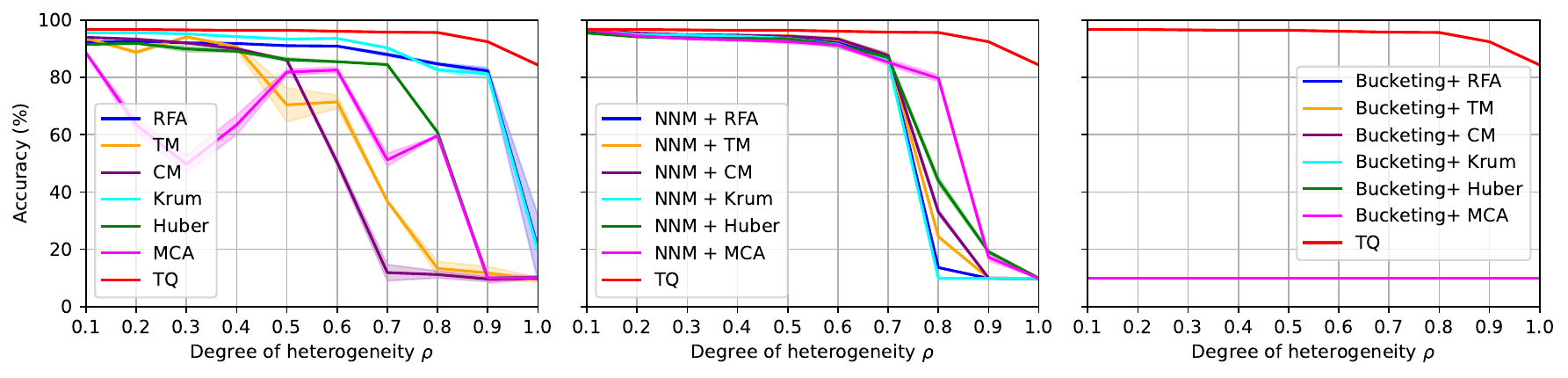}
\end{center}
\caption{Classification accuracy versus degree of heterogeneity for aggregators with or without pre-aggregation under $30\%$ Byzantine clients. Left: aggregators only; middle: aggregators with NNM; right: aggregators with Bucketing.}
\label{fig:degree-hene}
\end{figure*}

\begin{figure}[h]
\begin{center}
\includegraphics[width=0.95\linewidth]{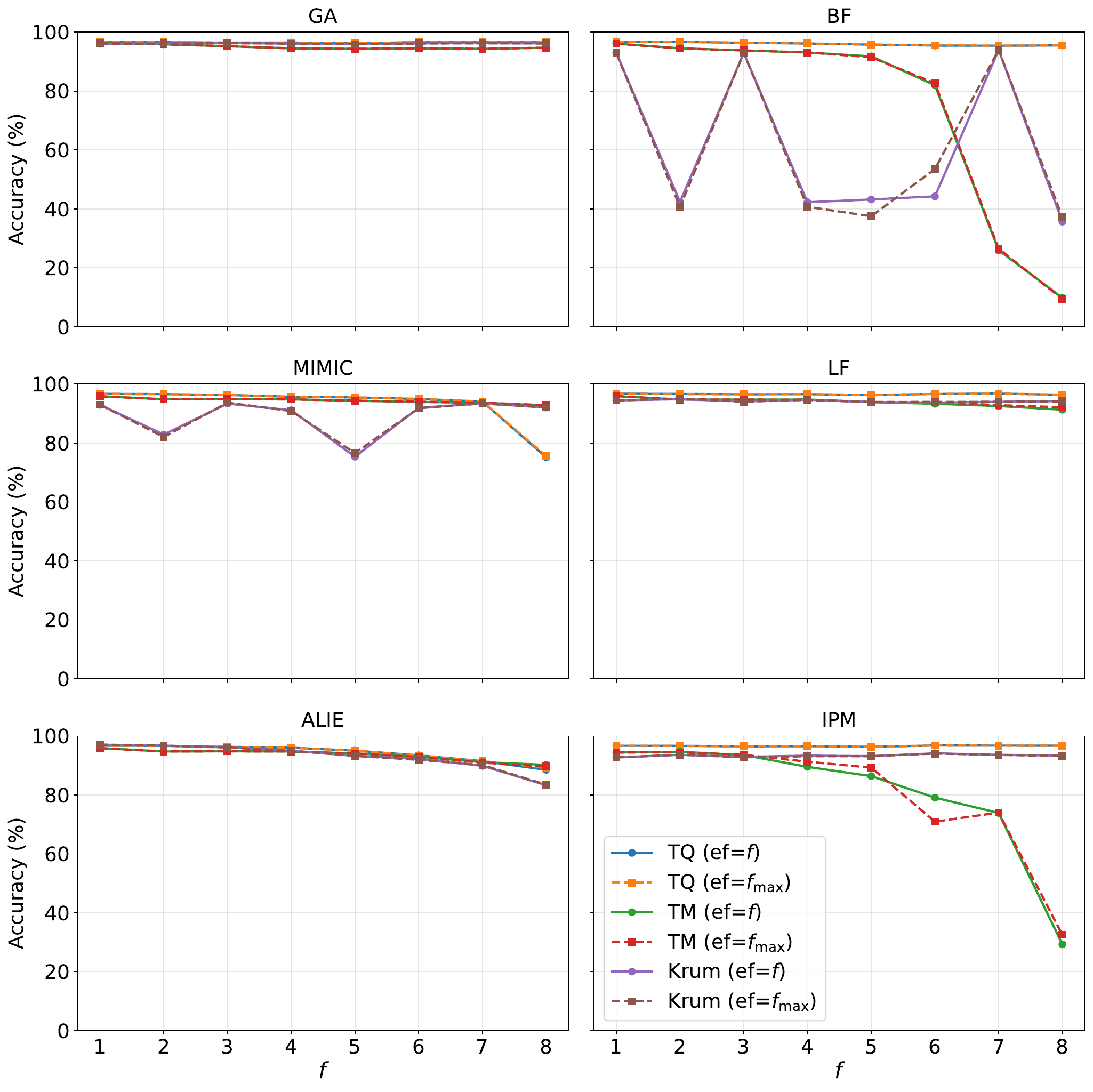}
\end{center}
\caption{Accuracy comparison under varying estimates of Byzantine clients for TQ, TM and Krum. Here, ef is the estimated number of Byzantine clients, ef=$f$ refers to that we input the true number of Byzantine clients to these aggregators, while ef=$f_{\max}$
refers to that $f_{\max}$ is considered as the estimated number of Byzantine clients.}
\label{fig:Aggregator_f_ef}
\end{figure}

\section{Experiments }\label{Sec:Ex_res}

\subsection{Experimental Settings}
\textbf{Datasets, heterogeneity and learning models.} Three datasets, i.e., MNIST \cite{lecun2010mnist}, Fashion-MNIST \cite{xiao2017fashion} and CIFAR-10 \cite{krizhevsky2009learning}, are adopted. 
 We follow the strategy in \cite{karimireddy2022byzantinerobust} to generate heterogeneous data. Denoting $0\leq\rho\leq 1$ as the degree of heterogeneity, a fraction $1-\rho$ of the training data is first assigned randomly and evenly across the $n$ clients. The remaining data are sorted by label and then evenly distributed to the clients in this order. A higher value of $\rho$ results in greater data heterogeneity.
The CNN architecture from \cite{karimireddy2022byzantinerobust} is used for MNIST and Fashion-MNIST, while ResNet-20 \cite{he2016deep} is adopted for CIFAR-10.

\textbf{Training settings, aggregators and malicious attacks.} The total numbers of clients for MNIST and CIFAR-10 are $n=20$ and $n=25$, respectively.
The parameters for MNIST in Algorithm \ref{Algo:ROGC} are $b=32$, $\mu=0.99$, $\lambda=0.02$ and $T=1500$. Fashion-MNIST has the same parameters as MNIST.
For CIFAR-10 we use $b=32$, $\mu=0.9$, $\lambda=0.1$ and $T=3000$, with a learning rate decay $0.1$ at $0.75\times T$. 
While for the NLP task, $b=32$, $\mu=0.9$, $\lambda=0.01$ and $T=900$.
The loss functions for MNIST, CIFAR-10 and the NLP task are negative log-likelihood loss, cross-entropy loss, cross-entropy loss, respectively.
Robust aggregation rules, including RFA~\cite{pillutla2022robust}, CM, TM~\cite{yin2018byzantine}, Krum~\cite{allouah2023fixing}, Huber~\cite{zhao2024huber}, MCA \cite{luan2024robust} and our TQ, are employed. 
RFA, Huber, MCA and TQ are iterative; we set the number of iterations to ten for each.
Furthermore, six different attacks are considered, including a little
is enough (ALIE)~\cite{baruch2019little}, label-flipping (LF)~\cite{allen2020byzantine}, bit-flipping (BF)~\cite{karimireddy2021learning}, inner product manipulation (IPM)~\cite{xie2020fall}, MIMIC~\cite{karimireddy2022byzantinerobust} and Gaussian attack (GA)~\cite{dong2023byzantine}.
Codes for aggregators and attacks are primarily adapted from \cite{karimireddy2021learning} and \cite{allouah2023fixing}.

It is worth noting that to make Krum provide a satisfactory
performance, its revised version in \cite{allouah2023fixing} is adopted.
{In addition, IPM sends the attack vector $\pmb a_t=-c_0 \cdot \overline{\pmb x}^t$ to the server, while \cite{karimireddy2022byzantinerobust} sets  $c_0=0.1$, yielding that the IPM attack is not an outlier and can directly handled by the arithmetic mean.
This study set $c_0=-3(n-f)/f$ to ensure that the magnitude of vector generated by the IPM exceeds that of the good gradients.
In comparison to the IPM, the magnitudes of the attack vectors given by MIMIC and ALIE attacks are smaller than those of the good vectors, which can be regarded as inlier noise. }

Furthermore, the experiments on MNIST, Fashion-MNIST, and CIFAR-10 were conducted on a Supermicro X11DPG-OT server with dual Intel Xeon Silver 4108 CPUs (1.80 GHz), 500 GB of RAM, a 2.13 TB SSD, and an NVIDIA GeForce RTX 2080 Ti GPU. 

\subsection{Experimental Results}

{We first investigate the effectiveness of the robust estimation strategy for the hyperparameter $\tau_k$ under varying numbers of Byzantine clients and degrees of data heterogeneity. The results on MNIST are presented in Fig. \ref{fig-MAD}. It is observed that TQ performs comparably to TQ$_t$ across different numbers of adversaries and levels of heterogeneity.} Following this, we conduct the following experiments under various scenarios.

\textbf{Impact of number of Byzantine clients on performance.}  
Fig.~\ref{fig:number-By} plots the classification performance versus number of Byzantine clients under the IPM attack on MNIST dataset. 
The left subfigure shows that TQ outperforms all competitors across the full range of adversaries. 
Pre-aggregation steps, including NNM and Bucketing, are employed to enhance the robustness of these competitors; their performance curves are shown in the middle and right subfigures of Fig.~\ref{fig:number-By}, respectively.
It is seen that Bucketing reduces the breakdown point of aggregators because Bucketing + aggregators fail when the number of Byzantine clients exceeds $4$, whereas the same aggregators can tolerate at least $6$ (i.e., $30\%$) without Bucketing. 
Besides, NNM can enhance robustness of aggregators under heterogeneity $\rho=0.5$ since the performance of aggregators degrades for $f>6$ in the left figure, while NNM + aggregators still maintain high accuracy even at $f=9$.
Nonetheless, TQ achieves the highest accuracy across all $f$ and remains effective even as the Byzantine fraction approaches $50\%$. 
Moreover, TQ outperforms the competing aggregators with NNM or Bucketing for all $f$, whereas the accuracy of Huber decreases with $f$ because Huber yields a biased solution and the bias increases with the fraction of Byzantine clients.
Furthermore, the performance of MCA is inferior to that of other aggregators because, although we set its hyperparameter based on \cite{luan2024robust}, it requires a proper adaptive hyperparameter to control the size of the Gaussian kernel.

\begin{table*}
\caption{Average accuracy (mean$\pm$standard deviation) at $\rho=0.5$ under different number of Byzantine clients on MNIST.}
\begin{center}
\setlength{\tabcolsep}{2mm}{
\begin{tabular}{cccccccc >{\columncolor{blue!5}} c}
\toprule 
&{Aggregator} & {GA} & {ALIE} & {BF} & {IPM} & {LF} & {MIMIC} & \textbf{Worst Case} \\
\hline
\multirow{7}{*}{\begin{tabular}[c]{@{}c@{}}$f=5$\end{tabular}}& RFA \cite{pillutla2022robust} & 96.46$\pm$0.17 & 93.40$\pm$0.32 & 21.61$\pm$17.75 & 92.63$\pm$0.79 & 96.82$\pm$0.36 & 95.31$\pm$1.12 & {21.61$\pm$17.75} \\
\cline{2-9} 
&\multirow{1}{*}{\begin{tabular}[c]{@{}c@{}}TM \cite{yin2018byzantine}\end{tabular}} & 94.31$\pm$0.33 & 94.10$\pm$0.63 & 91.60$\pm$0.93 & 87.92$\pm$5.87 & 93.79$\pm$0.38 & 94.34$\pm$0.32 & {87.92$\pm$5.87} \\
\cline{2-9} 
&\multirow{1}{*}{\begin{tabular}[c]{@{}c@{}}CM \cite{yin2018byzantine}\end{tabular}} & 94.86$\pm$0.28 & 93.36$\pm$0.39 & 93.63$\pm$0.19 & 88.08$\pm$0.88 & 95.04$\pm$0.10 & 93.57$\pm$0.74 & {88.08$\pm$0.88} \\
\cline{2-9} 
&\multirow{1}{*}{\begin{tabular}[c]{@{}c@{}}Krum \cite{blanchard2017machine}\end{tabular}} & 95.84$\pm$0.49 & 93.25$\pm$0.49 & 12.62$\pm$4.11 & 93.09$\pm$0.22 & 95.86$\pm$0.18 & 88.25$\pm$4.38 & {12.62$\pm$4.11} \\
\cline{2-9} 
&\multirow{1}{*}{\begin{tabular}[c]{@{}c@{}}Huber \cite{zhao2024huber}\end{tabular}} & 96.67$\pm$0.17 & 95.86$\pm$0.59 & 9.82$\pm$0.00 & 32.28$\pm$47.20 & 97.19$\pm$0.25 & 96.48$\pm$0.39 & {9.82$\pm$0.00} \\
\cline{2-9} 
&\multirow{1}{*}{\begin{tabular}[c]{@{}c@{}}MCA \cite{luan2024robust}\end{tabular}} & 96.69$\pm$0.16 & 95.97$\pm$0.56 & 13.11$\pm$3.05 & 76.57$\pm$10.45 & 97.10$\pm$0.66 & 96.55$\pm$0.31 & {13.11$\pm$3.05} \\
\cline{2-9} 
&\multirow{1}{*}{\begin{tabular}[c]{@{}c@{}}\textbf{TQ}\end{tabular}} & {96.15$\pm$0.10} & {95.03$\pm$0.25} & {95.76$\pm$0.42} & {96.32$\pm$0.18} & 96.31$\pm$0.34 & {95.45$\pm$0.74} & {\bf{95.03$\pm$0.25}} \\
\hline
\multirow{7}{*}{\begin{tabular}[c]{@{}c@{}}$f=6$\end{tabular}}&\multirow{1}{*}{\begin{tabular}[c]{@{}c@{}}RFA \cite{pillutla2022robust}\end{tabular}} & 96.34$\pm$0.12 & 92.31$\pm$0.42 & 9.55$\pm$0.59 & 91.07$\pm$0.82 & {97.16$\pm$0.28} & 94.24$\pm$0.83 & {9.55$\pm$0.59} \\
\cline{2-9} 
&\multirow{1}{*}{\begin{tabular}[c]{@{}c@{}}TM \cite{yin2018byzantine}\end{tabular}} & 94.53$\pm$0.43 & 92.77$\pm$0.50 & 82.72$\pm$1.04 & 75.35$\pm$24.70 & 92.91$\pm$0.56 & 94.01$\pm$0.26 & {75.35$\pm$24.70} \\
\cline{2-9} 
&\multirow{1}{*}{\begin{tabular}[c]{@{}c@{}}CM \cite{yin2018byzantine}\end{tabular}} & 94.60$\pm$0.18 & 92.66$\pm$0.59 & 91.12$\pm$1.00 & 86.80$\pm$2.26 & 94.57$\pm$0.41 & 92.37$\pm$0.37 & {86.80$\pm$2.26} \\
\cline{2-9} 
&\multirow{1}{*}{\begin{tabular}[c]{@{}c@{}}Krum \cite{blanchard2017machine}\end{tabular}} & 96.19$\pm$0.11 & 92.28$\pm$0.16 & 50.30$\pm$40.21 & 93.84$\pm$0.67 & 93.98$\pm$1.18 & 91.27$\pm$1.24 & {50.30$\pm$40.21} \\
\cline{2-9} 
&\multirow{1}{*}{\begin{tabular}[c]{@{}c@{}}Huber \cite{zhao2024huber}\end{tabular}} & 96.64$\pm$0.13 & 95.11$\pm$0.42 & 9.82$\pm$0.00 & 80.17$\pm$4.74 & 96.48$\pm$0.29 & 96.26$\pm$0.41  & {9.82$\pm$0.00} \\
\cline{2-9} 
&\multirow{1}{*}{\begin{tabular}[c]{@{}c@{}}MCA \cite{luan2024robust}\end{tabular}} & 96.58$\pm$0.10 & 95.44$\pm$0.30 & 10.95$\pm$0.72 & 81.93$\pm$7.62 & 96.32$\pm$0.70 & 96.41$\pm$0.27  & {10.95$\pm$0.72} \\
\cline{2-9} 
&\multirow{1}{*}{\begin{tabular}[c]{@{}c@{}}\textbf{TQ}\end{tabular}} & {96.59$\pm$0.06} & {93.42$\pm$0.59} & {95.39$\pm$0.52} & {96.38$\pm$0.32} & 96.63$\pm$0.42 & {94.95$\pm$1.30} & {\bf{93.42$\pm$0.59}} \\
\hline
\multirow{7}{*}{\begin{tabular}[c]{@{}c@{}}$f=7$\end{tabular}}&\multirow{1}{*}{\begin{tabular}[c]{@{}c@{}}RFA \cite{pillutla2022robust}\end{tabular}} & 96.39$\pm$0.18 & 90.37$\pm$0.25 & 9.54$\pm$0.52 & 89.42$\pm$0.85 & 96.45$\pm$0.71 & 94.13$\pm$0.89 & {9.54$\pm$0.52} \\
\cline{2-9} 
&\multirow{1}{*}{\begin{tabular}[c]{@{}c@{}}TM \cite{yin2018byzantine}\end{tabular}} & 94.31$\pm$0.36 & 91.23$\pm$0.18 & 31.69$\pm$23.79 & 60.57$\pm$15.36 & 92.40$\pm$0.90 & 93.68$\pm$0.59 & {31.69$\pm$23.79} \\
\cline{2-9} 
&\multirow{1}{*}{\begin{tabular}[c]{@{}c@{}}CM \cite{yin2018byzantine}\end{tabular}} & 94.41$\pm$0.24 & 91.94$\pm$0.38 & 83.61$\pm$3.42 & 57.84$\pm$11.53 & 92.87$\pm$0.56 & 92.11$\pm$0.94 & {57.84$\pm$11.53} \\
\cline{2-9} 
&\multirow{1}{*}{\begin{tabular}[c]{@{}c@{}}Krum \cite{blanchard2017machine}\end{tabular}} & 96.29$\pm$0.17 & 90.03$\pm$0.53 & 66.01$\pm$48.55 & 93.37$\pm$0.51 & 95.99$\pm$0.36 & 65.00$\pm$47.88 & {66.01$\pm$48.55} \\
\cline{2-9} 
&\multirow{1}{*}{\begin{tabular}[c]{@{}c@{}}Huber \cite{zhao2024huber}\end{tabular}} & 96.63$\pm$0.11 & 94.33$\pm$0.15 & 9.82$\pm$0.00 & 81.45$\pm$1.31 & 86.87$\pm$6.52 & 96.02$\pm$0.18 & {9.82$\pm$0.00} \\
\cline{2-9} 
&\multirow{1}{*}{\begin{tabular}[c]{@{}c@{}}MCA \cite{luan2024robust}\end{tabular}} & 96.67$\pm$0.12 & 94.46$\pm$0.07 & 11.01$\pm$0.62 & 12.85$\pm$3.79 & 89.49$\pm$4.60 & 96.11$\pm$0.18 & {11.01$\pm$0.62} \\
\cline{2-9} 
&\multirow{1}{*}{\begin{tabular}[c]{@{}c@{}}\textbf{TQ}\end{tabular}} & {96.63$\pm$0.12} & {91.54$\pm$0.51} & {95.40$\pm$0.33} & {96.48$\pm$0.41} & 96.75$\pm$0.22 & {94.05$\pm$0.46} & {\bf{91.54$\pm$0.51}} \\
\bottomrule 
\end{tabular}}
\label{table:agg}
\end{center}
\end{table*}

\begin{table*}
\caption{Average accuracy (mean$\pm$standard deviation) at $f=6$ under varying levels of heterogeneity on MNIST.}
\begin{center}
\setlength{\tabcolsep}{2mm}{
\begin{tabular}{cccccccc >{\columncolor{blue!5}} c}
\toprule 
&{Aggregator} & {GA} & {ALIE} & {BF} & {IPM} & {LF} & {MIMIC} & \textbf{Worst Case} \\
\hline
\multirow{7}{*}{\begin{tabular}[c]{@{}c@{}}$\rho=0.6$\end{tabular}}& \multirow{1}{*}{\begin{tabular}[c]{@{}c@{}}RFA \cite{pillutla2022robust}\end{tabular}} & 96.25$\pm$0.30 & 90.67$\pm$0.61 & 9.15$\pm$0.38 & 90.91$\pm$0.68 & 96.70$\pm$0.03 & 93.03$\pm$0.71 & {9.15$\pm$0.38} \\
\cline{2-9} 
&\multirow{1}{*}{\begin{tabular}[c]{@{}c@{}}TM \cite{yin2018byzantine}\end{tabular}} & 93.93$\pm$0.39 & 91.57$\pm$0.29 & 33.62$\pm$16.15 & 74.99$\pm$7.11 & 91.78$\pm$1.20 & 93.20$\pm$0.14 & {33.62$\pm$16.15} \\
\cline{2-9} 
&\multirow{1}{*}{\begin{tabular}[c]{@{}c@{}}CM \cite{yin2018byzantine}\end{tabular}} & 94.03$\pm$0.27 & 90.92$\pm$0.10 & 75.67$\pm$0.63 & 52.77$\pm$33.67 & 92.87$\pm$0.71 & 90.56$\pm$0.55 & {52.77$\pm$33.67} \\
\cline{2-9} 
&\multirow{1}{*}{\begin{tabular}[c]{@{}c@{}}Krum \cite{blanchard2017machine}\end{tabular}} & 95.72$\pm$0.53 & 90.30$\pm$0.73 & 10.44$\pm$0.82 & 93.61$\pm$0.47 & 94.85$\pm$0.92 & 71.34$\pm$37.45 & {10.44$\pm$0.82} \\
\cline{2-9} 
&\multirow{1}{*}{\begin{tabular}[c]{@{}c@{}}Huber \cite{zhao2024huber}\end{tabular}} & 96.70$\pm$0.26 & 94.58$\pm$0.88 & 9.82$\pm$0.00 & 81.05$\pm$2.66 & 94.70$\pm$1.20 & 96.11$\pm$0.42 & {9.82$\pm$0.00} \\
\cline{2-9} 
&\multirow{1}{*}{\begin{tabular}[c]{@{}c@{}}MCA \cite{luan2024robust}\end{tabular}} & 96.67$\pm$0.25 & 95.01$\pm$0.79 & 12.22$\pm$2.93 & 82.59$\pm$1.51 & 92.89$\pm$2.30 & 96.36$\pm$0.31 & {12.22$\pm$2.93} \\
\cline{2-9} 
&\multirow{1}{*}{\begin{tabular}[c]{@{}c@{}}\textbf{TQ}\end{tabular}} & {96.51$\pm$0.27} & {92.46$\pm$0.60} & {94.59$\pm$0.55} & {96.11$\pm$0.59} & 96.35$\pm$0.55 & {94.02$\pm$0.95} & {\bf{92.46$\pm$0.60}} \\
\hline
\multirow{7}{*}{\begin{tabular}[c]{@{}c@{}}$\rho=0.7$\end{tabular}}&\multirow{1}{*}{\begin{tabular}[c]{@{}c@{}}RFA \cite{pillutla2022robust}\end{tabular}} & 96.12$\pm$0.16 & 89.82$\pm$0.33 & 9.39$\pm$0.79 & 88.01$\pm$2.55 & 96.10$\pm$0.06 & 91.14$\pm$1.77 & {9.39$\pm$0.79} \\
\cline{2-9} 
&\multirow{1}{*}{\begin{tabular}[c]{@{}c@{}}TM \cite{yin2018byzantine}\end{tabular}} & 92.97$\pm$0.40 & 89.69$\pm$0.87 & 10.29$\pm$0.43 & 15.47$\pm$5.28 & 88.08$\pm$1.83 & 91.90$\pm$1.14 & {10.29$\pm$0.43} \\
\cline{2-9} 
&\multirow{1}{*}{\begin{tabular}[c]{@{}c@{}}CM \cite{yin2018byzantine}\end{tabular}} & 92.66$\pm$0.65 & 90.07$\pm$0.32 & 35.98$\pm$21.33 & 9.21$\pm$0.84 & 90.22$\pm$2.43 & 91.27$\pm$1.64 & {9.21$\pm$0.84} \\
\cline{2-9} 
&\multirow{1}{*}{\begin{tabular}[c]{@{}c@{}}Krum \cite{blanchard2017machine}\end{tabular}} & 94.85$\pm$1.05 & 89.25$\pm$0.59 & 10.21$\pm$1.07 & 90.30$\pm$0.97 & 92.99$\pm$0.86 & 60.11$\pm$43.06 & {10.21$\pm$1.07} \\
\cline{2-9} 
&\multirow{1}{*}{\begin{tabular}[c]{@{}c@{}}Huber \cite{zhao2024huber}\end{tabular}} & 96.67$\pm$0.10 & 93.99$\pm$0.86 & 9.82$\pm$0.00 & 82.33$\pm$2.30 & 91.17$\pm$6.18 & 95.67$\pm$0.48 & {9.82$\pm$0.00} \\
\cline{2-9} 
&\multirow{1}{*}{\begin{tabular}[c]{@{}c@{}}MCA \cite{luan2024robust}\end{tabular}} & 96.64$\pm$0.16 & 94.17$\pm$1.07 & 9.97$\pm$1.26 & 51.21$\pm$31.34 & 89.52$\pm$4.17 & 95.94$\pm$0.39 & {9.97$\pm$1.26} \\
\cline{2-9} 
&\multirow{1}{*}{\begin{tabular}[c]{@{}c@{}}\textbf{TQ}\end{tabular}} & {96.18$\pm$0.59} & {92.07$\pm$0.90} & {93.83$\pm$1.42} & {95.81$\pm$0.20} & 96.46$\pm$0.59 & {92.64$\pm$0.84} & {\bf{92.07$\pm$0.90}} \\
\bottomrule 
\end{tabular}}
\label{table:agg_hetero}
\end{center}
\end{table*}

\begin{table*}
\caption{Average accuracy (mean$\pm$standard deviation) at $\rho=0.4$ under different attacks on CIFAR-10.}
\begin{center}
\setlength{\tabcolsep}{2.2mm}{
\begin{tabular}{cccccccc >{\columncolor{blue!5}} c}
\toprule 
&{Aggregator} & {GA} & {ALIE} & {BF} & {IPM} & {LF} & {MIMIC} & \textbf{Worst Case} \\
\hline
\multirow{7}{*}{\begin{tabular}[c]{@{}c@{}}$f=2$\end{tabular}}&\multirow{1}{*}{\begin{tabular}[c]{@{}c@{}}RFA \cite{pillutla2022robust}\end{tabular}} & 72.04$\pm$0.11 & 69.89$\pm$0.33 & 68.61$\pm$0.20 & 58.83$\pm$0.99 & 70.73$\pm$0.18 & 71.75$\pm$0.88 & {58.83$\pm$0.99} \\
\cline{2-9} 
&\multirow{1}{*}{\begin{tabular}[c]{@{}c@{}}TM \cite{yin2018byzantine}\end{tabular}} & 71.09$\pm$0.26 & 69.77$\pm$0.48 & 66.29$\pm$0.72 & 56.02$\pm$0.44 & 67.36$\pm$0.75 & 69.38$\pm$0.48 & {56.02$\pm$0.44} \\
\cline{2-9} 
&\multirow{1}{*}{\begin{tabular}[c]{@{}c@{}}CM \cite{yin2018byzantine}\end{tabular}} & 67.22$\pm$0.68 & 67.50$\pm$0.17 & 64.77$\pm$1.35 & 59.94$\pm$0.72 & 65.88$\pm$1.21 & 65.87$\pm$0.60 & {59.94$\pm$0.72} \\
\cline{2-9} 
&\multirow{1}{*}{\begin{tabular}[c]{@{}c@{}}Krum \cite{blanchard2017machine}\end{tabular}} & 52.62$\pm$0.62 & 68.63$\pm$0.59 & 35.06$\pm$2.26 & 42.15$\pm$1.86 & 47.12$\pm$2.33 & 36.35$\pm$1.92 & {35.06$\pm$2.26} \\
\cline{2-9} 
&\multirow{1}{*}{\begin{tabular}[c]{@{}c@{}}Huber \cite{zhao2024huber}\end{tabular}} & 71.99$\pm$0.35 & 71.64$\pm$0.66 & 66.89$\pm$0.09 & 11.31$\pm$2.41 & 69.26$\pm$0.83 & 71.75$\pm$0.42 & {11.31$\pm$2.41} \\
\cline{2-9} 
&\multirow{1}{*}{\begin{tabular}[c]{@{}c@{}}MCA \cite{luan2024robust}\end{tabular}} & 71.18$\pm$0.11 & 71.83$\pm$0.43 & 67.39$\pm$0.67 & 41.29$\pm$2.41 & 70.10$\pm$0.29 & 71.51$\pm$0.69 & {41.29$\pm$2.41} \\
\cline{2-9} 
&\multirow{1}{*}{\begin{tabular}[c]{@{}c@{}}\textbf{TQ}\end{tabular}} & 71.55$\pm$0.28 & 71.62$\pm$0.70 & 67.40$\pm$0.27 & 71.57$\pm$0.28 & 71.67$\pm$0.53 & 71.27$\pm$0.26 & {\bf{67.40$\pm$0.27}} \\
\hline
\multirow{7}{*}{\begin{tabular}[c]{@{}c@{}}$f=3$\end{tabular}}&\multirow{1}{*}{\begin{tabular}[c]{@{}c@{}}RFA \cite{pillutla2022robust}\end{tabular}} & 72.30$\pm$0.76 & 64.56$\pm$1.19 & 66.44$\pm$0.50 & 56.81$\pm$0.49 & 70.33$\pm$0.72 & 70.63$\pm$0.33 & {64.56$\pm$1.19} \\
\cline{2-9} 
&\multirow{1}{*}{\begin{tabular}[c]{@{}c@{}}TM \cite{yin2018byzantine}\end{tabular}} & 70.30$\pm$0.97 & 67.75$\pm$0.83 & 63.84$\pm$1.16 & 50.85$\pm$1.19 & 65.77$\pm$0.92 & 67.10$\pm$1.13 & {63.84$\pm$1.16} \\
\cline{2-9} 
&\multirow{1}{*}{\begin{tabular}[c]{@{}c@{}}CM \cite{yin2018byzantine}\end{tabular}} & 66.58$\pm$0.88 & 64.61$\pm$0.81 & 64.47$\pm$0.62 & 56.41$\pm$0.54 & 65.33$\pm$0.78 & 65.41$\pm$0.57 & {56.41$\pm$0.54} \\
\cline{2-9} 
&\multirow{1}{*}{\begin{tabular}[c]{@{}c@{}}Krum \cite{blanchard2017machine}\end{tabular}} & 51.92$\pm$2.06 & 60.53$\pm$0.61 & 42.62$\pm$0.99 & 41.30$\pm$1.60 & 47.41$\pm$1.55 & 45.99$\pm$0.47 & {42.62$\pm$0.99} \\
\cline{2-9} 
&\multirow{1}{*}{\begin{tabular}[c]{@{}c@{}}Huber \cite{zhao2024huber}\end{tabular}} & 71.47$\pm$0.73 & 71.75$\pm$0.86 & 63.60$\pm$1.02 & 35.79$\pm$1.92 & 68.72$\pm$0.92 & 71.32$\pm$0.56 & {35.79$\pm$1.92} \\
\cline{2-9} 
&\multirow{1}{*}{\begin{tabular}[c]{@{}c@{}}MCA \cite{luan2024robust}\end{tabular}} & 71.83$\pm$0.35 & 71.78$\pm$0.67 & 63.91$\pm$0.33 & 37.72$\pm$1.97 & 68.60$\pm$0.63 & 70.94$\pm$0.53 & {37.72$\pm$1.97} \\
\cline{2-9} 
&\multirow{1}{*}{\begin{tabular}[c]{@{}c@{}}\textbf{TQ}\end{tabular}} & {71.74$\pm$0.48} & {71.43$\pm$1.07} & {67.68$\pm$0.79} & {72.00$\pm$0.58} & 72.17$\pm$1.12 & {70.91$\pm$0.73} & {\bf{67.68$\pm$0.79}} \\
\hline
\multirow{7}{*}{\begin{tabular}[c]{@{}c@{}}$f=4$\end{tabular}}&\multirow{1}{*}{\begin{tabular}[c]{@{}c@{}}RFA \cite{pillutla2022robust}\end{tabular}} & 72.09$\pm$1.44 & 53.33$\pm$2.53 & 62.30$\pm$1.32 & 54.65$\pm$1.13 & 68.70$\pm$0.75 & 69.47$\pm$0.45 & {53.33$\pm$2.53} \\
\cline{2-9} 
&\multirow{1}{*}{\begin{tabular}[c]{@{}c@{}}TM \cite{yin2018byzantine}\end{tabular}} & 69.52$\pm$0.16 & 66.72$\pm$1.05 & 61.97$\pm$0.73 & 46.76$\pm$2.41 & 65.61$\pm$0.83 & 67.01$\pm$1.17 & {61.97$\pm$0.73} \\
\cline{2-9} 
&\multirow{1}{*}{\begin{tabular}[c]{@{}c@{}}CM \cite{yin2018byzantine}\end{tabular}} & 66.80$\pm$0.02 & 57.79$\pm$0.65 & 63.38$\pm$0.42 & 51.23$\pm$0.18 & 65.29$\pm$0.83 & 64.23$\pm$1.32 & {51.23$\pm$0.18} \\
\cline{2-9} 
&\multirow{1}{*}{\begin{tabular}[c]{@{}c@{}}Krum \cite{blanchard2017machine}\end{tabular}} & 51.18$\pm$1.23 & 39.86$\pm$6.39 & 41.52$\pm$3.93 & 46.03$\pm$0.83 & 50.82$\pm$0.77 & 40.08$\pm$3.13 & {39.86$\pm$6.39} \\
\cline{2-9} 
&\multirow{1}{*}{\begin{tabular}[c]{@{}c@{}}Huber \cite{zhao2024huber}\end{tabular}} & 71.86$\pm$1.46 & 70.74$\pm$0.84 & 59.76$\pm$1.13 & 42.19$\pm$1.60 & 67.13$\pm$0.96 & 71.09$\pm$0.20 & {42.19$\pm$1.60} \\
\cline{2-9} 
&\multirow{1}{*}{\begin{tabular}[c]{@{}c@{}}MCA \cite{luan2024robust}\end{tabular}} & 71.98$\pm$1.01 & 70.72$\pm$0.38 & 59.36$\pm$1.61 & 32.19$\pm$0.72 & 66.80$\pm$1.01 & 70.94$\pm$0.17 & {32.19$\pm$0.72} \\
\cline{2-9} 
&\multirow{1}{*}{\begin{tabular}[c]{@{}c@{}}\textbf{TQ}\end{tabular}} & {71.79$\pm$0.94} & {69.10$\pm$0.97} & {68.44$\pm$1.19} & {68.24$\pm$0.90} & 71.50$\pm$0.99 & {70.89$\pm$0.55} & {\bf{68.44$\pm$1.19}} \\
\bottomrule 
\end{tabular}}
\label{table:agg-cifar}
\end{center}
\end{table*}

\begin{figure*}[h]
\begin{center}
\includegraphics[width=0.99\linewidth]{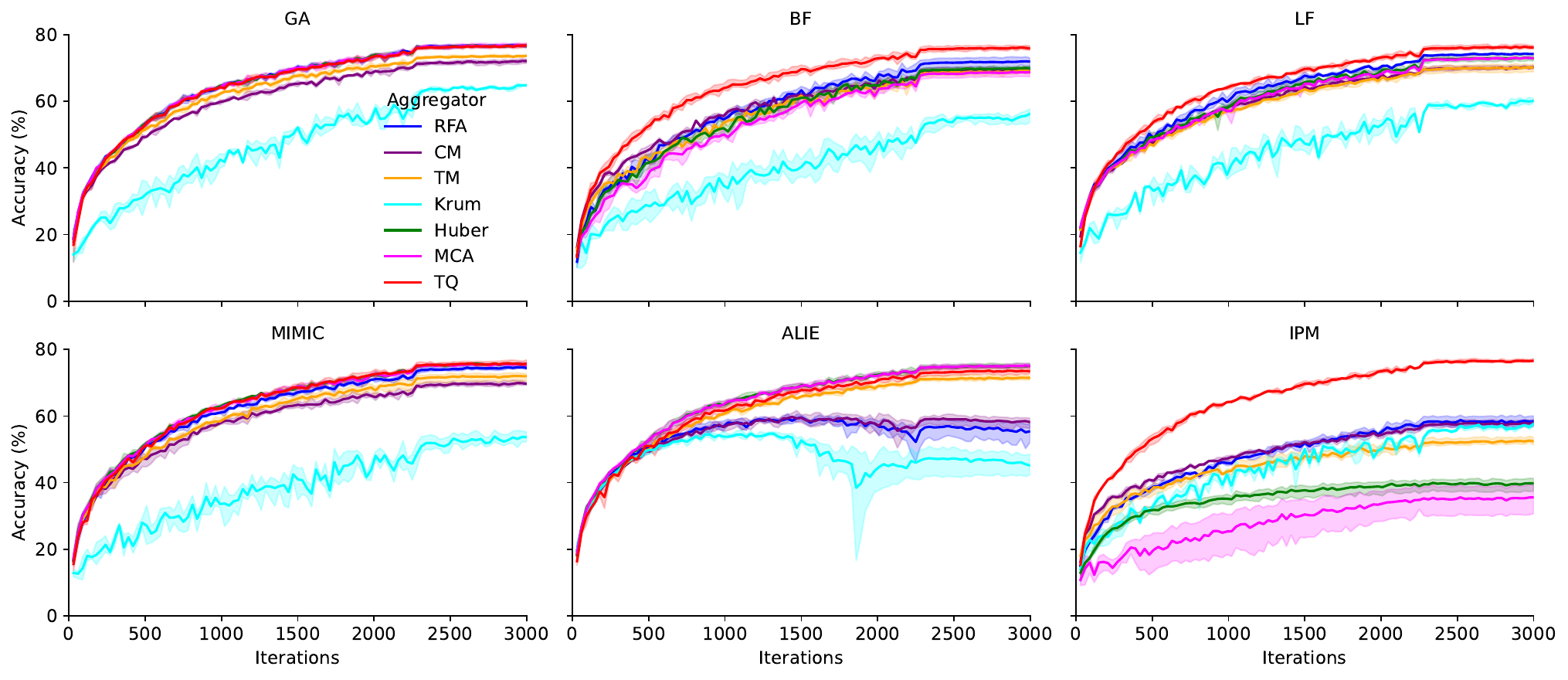}
\end{center}
\caption{Experimental results under different attacks with $\rho=0.3$ and $f=4$ on CIFAR-10.}
\label{fig:cifar-noniid0.3-f4}
\end{figure*}

\begin{figure*}[!htp]
\begin{center}
\includegraphics[width=0.99\linewidth]{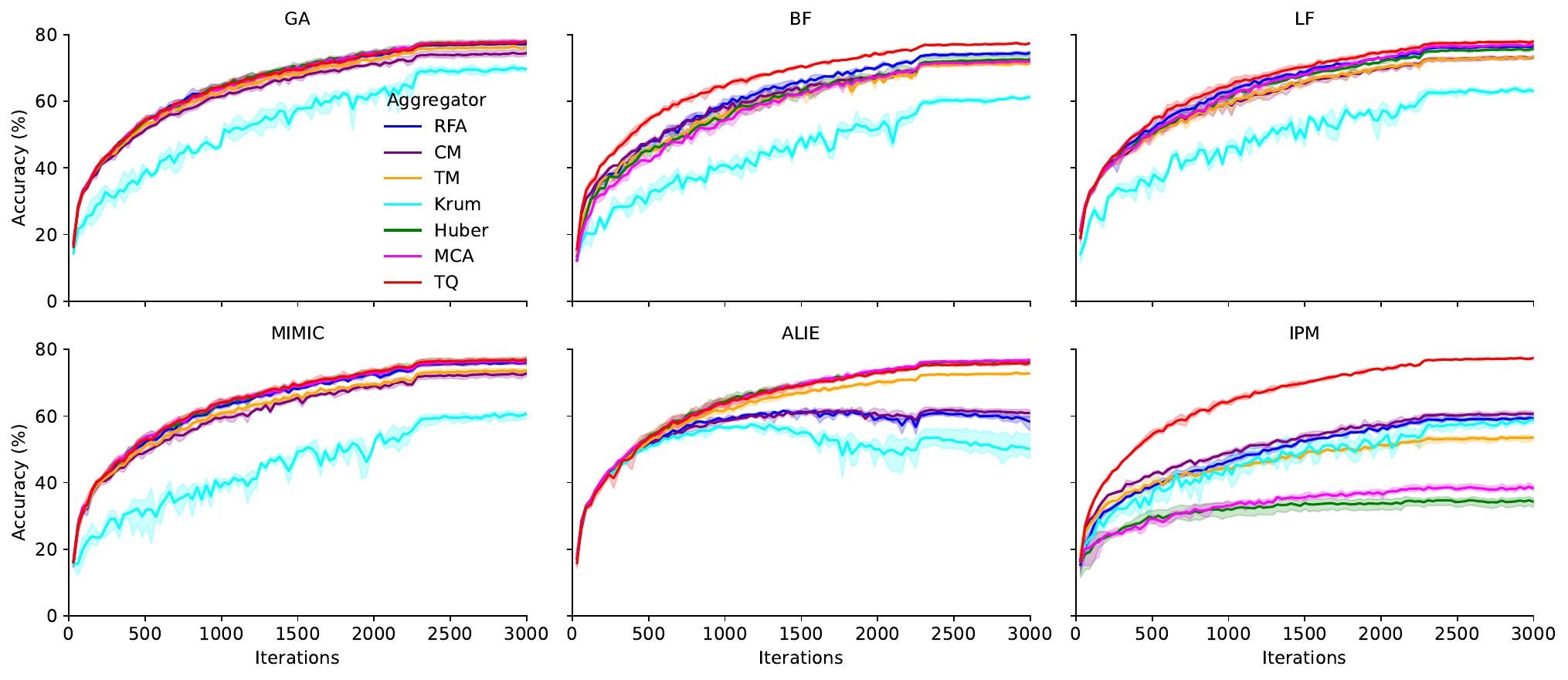}
\end{center}
\caption{Experimental results under different attacks with $\rho=0.2$ and $f=4$ on CIFAR-10.}
\label{fig:cifar-noniid0.2-f4}
\end{figure*}

\begin{figure*}[!htp]
\begin{center}
\includegraphics[width=0.99\linewidth]{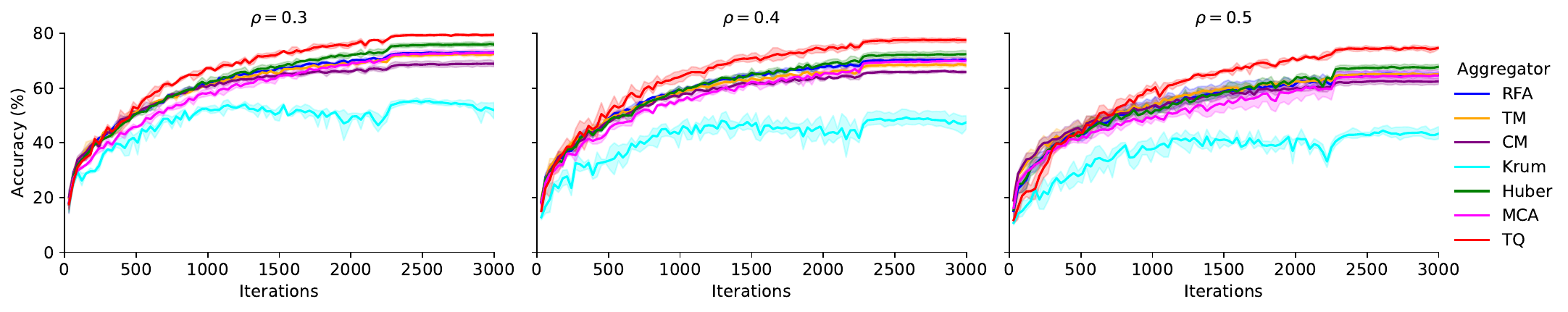}
\end{center}
\caption{Experimental results for different data heterogeneity under $f=6$ where the six attacks, namely, GA, BF, LF, MIMIC, ALIE and IPM, are adopted to attack the learning model at the same time.}
\label{fig:Cifar10-0.30.40.5-f6}
\end{figure*}

\begin{figure*}[!htp]
\begin{center}
\includegraphics[width=0.99\linewidth]{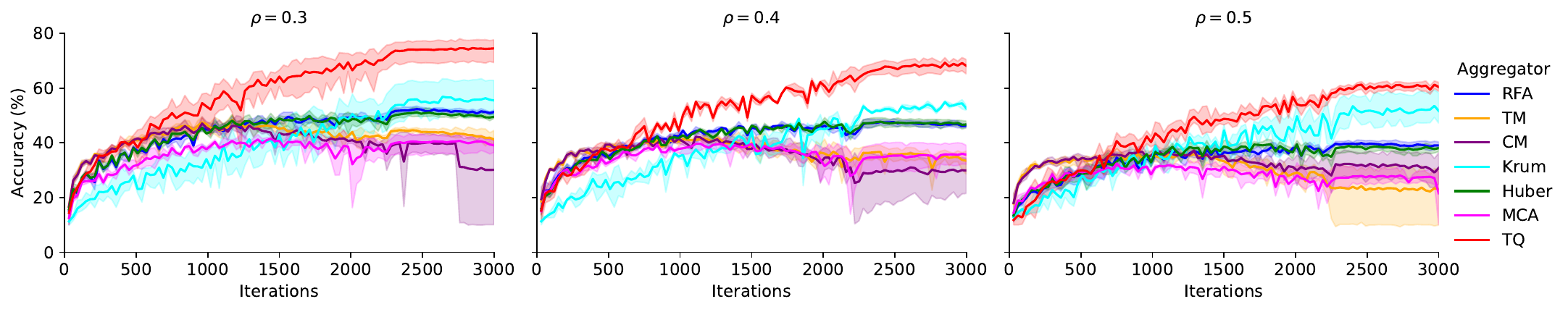}
\end{center}
\caption{Experimental results for different data heterogeneity under $f=12$ where the twelve attacks, namely, $2\times$GA, $2\times$BF, $2\times$LF, $2\times$MIMIC, $2\times$ALIE and $2\times$IPM attacks, are adopted to attack the learning model at the same time. Here, $2\times$GA attack refers to two GA attacks.}
\label{fig:Cifar10-0.30.40.5-f12}
\end{figure*}

\textbf{Impact of degree of heterogeneity on performance.}
Fig.~\ref{fig:degree-hene} plots the classification accuracy versus degree of heterogeneity under $30\%$ IPM attacks on the MNIST dataset.
As the percentage of Byzantine clients is bigger than $25\%$, it is shown that the combination of Bucketing plus aggregators fail in the right figure. 
The left figure shows that the performance of all aggregation rules degrades as the degree of heterogeneity of data increases. For example, TM and CM fail for $\rho \geq 0.8$. 
Besides, it is seen that NNM can enhance the robustness of aggregators such as RFA, TM, CM and Huber when $\rho\leq 0.6$ compared with their accuracies in the left figure, but their performance degrades with the growth of $\rho$, and these aggregation rules fail for $\rho > 0.8$. 
Since NNM averages the $n-f$ nearest neighbors of each vector (including itself), the $n-f$ nearest neighbors are easy to involve the gradients sent by Byzantine clients as $\rho$ increases, resulting in performance degradation.
However, TQ is superior to the competing robust rules without or with pre-aggregation and has the best classification accuracy for various degrees of data heterogeneity.

\textbf{Performance of aggregators for different attacks.}
Table~\ref{table:agg} tabulates the average classification results at $\rho=0.5$ under different numbers of Byzantine clients on MNIST dataset.
We see that all aggregators have comparable performance  
under the GA, ALIE, LF and MIMIC attacks for different numbers of malicious clients. 
While TQ significantly outperforms RFA, TM, CM, Huber and MCA under the IPM attack.
Besides, since each aggregator may encounter different malicious attacks simultaneously, the worst case for each rule under various attacks is adapted to measure the performance. 
It is seen that the worst accuracy for TQ is much higher than that of the remaining rules.
Moreover, we have evaluated all aggregators under varying levels of data heterogeneity, with the results presented in Table \ref{table:agg_hetero}. By combining the results at $\rho=0.5$ and $f=6$ from Table \ref{table:agg} with all findings in Table \ref{table:agg_hetero}, we observe that performance decreases as heterogeneity increases. Nevertheless, TQ achieves the highest accuracy in the worst-case scenario.
{Furthermore, since TQ, TM and Krum require estimating the number of Byzantine clients,
it is necessary to investigate the impact of $f$ on their performance.
Even though the true value of $f$ is unknown, setting the estimated value as $f_{\max}$ and $f_{\max}=\lfloor \frac{n-1}{2} \rfloor$ are reasonable.
For example, if $n=13$, we can provide $f_{\max}=6$ as the estimated value of $f$ for TQ, TM and Krum.
For each integrator, we provide the true value of $f$ and the estimated value $f_{\max}$, with the comparison results displayed in Fig. \ref{fig:Aggregator_f_ef}.
It is seen that although TM and Krum may yield poor results for certain attacks, the accuracy of TQ achieved by setting ef=$f_{\max}$ is comparable to that of ef=$f$ across all aggregators under various attacks.}
Extensive results on Fashion-MNIST are found in Appendix~G of the supplementary material.

\textbf{The performance of all aggregators is investigated on CIFAR-10.} 
Table~\ref{table:agg-cifar} tabulates the classification results at $\rho=0.4$ under different numbers of Byzantine clients.
Again, the worst accuracy of TQ for all attacks is higher than that of the remaining aggregators.
{Specially, it is seen that Huber and MCA achieve almost the highest accuracy for the MIMIC and ALIE attacks, while TQ shows comparable performance, since similar to Huber, TQ employs the quadratic function to handle the inlier noise.}
Besides, we observe that the performance of TQ is superior to that of the competitors when combating the BF, LF and IPM attacks for large number of malicious clients.
In contrast, Huber and MCA yield the worst classification accuracy for the IPM  attack, as when Huber handles outliers, it will result in bias, and the bias accumulates with communication rounds, while MCA needs a proper Gaussian kernel size.
Furthermore, the performance comparison of aggregators under varying data heterogeneity has been investigated and the results are shown in Figs. \ref{fig:cifar-noniid0.3-f4} and \ref{fig:cifar-noniid0.2-f4}.
By combining the results at $\rho=0.4$ and $f=4$ from Table \ref{table:agg-cifar}, we observe that TQ, MCA, Huber and RFA achieve comparable performance for GA and MIMIC attacks under varying levels of heterogeneity, and they are superior to CM, TM and Krum.
Nevertheless, the proposed aggregation rule outperforms the competitors for various values of $\rho$ under the BF, LF and IPM attacks.

\textbf{Performance comparison under mixed attacks.}
The above experiments consider that the learning process is attacked by the same attacks, that is, the learning algorithm is attacked by one of the GA, BF, LF, MIMIC, ALIE and IPM attacks but with varying number of attacks.
However, the real-world learning algorithm may be disturbed by different attacks at the same time. That is to say, if there exist six Byzantine clients, then it is possible that different Byzantine clients send different attacks, thus the received gradients sent by clients involve six different attacks.

To model the above scenario, we consider the following two cases. 
Case 1: The number of Byzantine clients is $6$ and these clients send six different attacks to the server under varying data heterogeneity, namely, $\rho = 0.3,~0.4,~0.5$.
Case 2: The number of Byzantine clients is $12$ and these clients send six different attacks to the server and the number of each attack is $2$ under varying data heterogeneity, namely, $\rho = 0.3,~0.4,~0.5$. 
Since the total number of clients is $n =25$, we only consider the two cases where the maximum Byzantine client number is $f_{\max}=\lfloor \frac{n-1}{2} \rfloor=12$.
Figs. \ref{fig:Cifar10-0.30.40.5-f6} and \ref{fig:Cifar10-0.30.40.5-f12} plot the results of Cases 1 and 2, respectively. 
It is seen that the accuracy of each aggregator decreases as the degree of data heterogeneity, i.e., $\rho$, increases for $f=6$ and $f=12$. 
Besides, increasing $f$ from $6$ to $12$ degrades the performance of all robust rules.
Nonetheless, TQ significantly outperforms RFA, TM, CM, Krum, Huber and MCA for different values of $\rho$ and $f$.

\section{Conclusion}\label{Sec:con}
In this paper, we demonstrate the equivalence of Huber and CC in terms of robustness using convex conjugate theory, and analyze how both robust rules introduce bias when handling outliers. 
This bias increases with the number of Byzantine clients and data heterogeneity. 
The underlying reason for this phenomenon is that Huber employs the $\ell_1$-norm to resist outliers. 
To address this issue, we propose a new aggregator called TQ, which clips all outliers.
We employ the robust dispersion to estimate the hyperparameter involved in our aggregator and analyze its effectiveness.
Additionally, we identify its equivalent form and analyze its compliance with $(f, \kappa)$-robustness. 
The robustness coefficient of TQ is optimal with respect to the fraction of Byzantine clients, enabling TQ to achieve order-optimal Byzantine-robust learning.
Experimental results demonstrate the superiority of TQ over the competing aggregators.

\ifCLASSOPTIONcaptionsoff
  \newpage
\fi

\bibliographystyle{IEEEtran}

\bibliography{Ref}

@inproceedings{mcmahan2017communication,
  title={Communication-efficient learning of deep networks from decentralized data},
  author={McMahan, Brendan and Moore, Eider and Ramage, Daniel and Hampson, Seth and y Arcas, Blaise Aguera},
  booktitle={Artificial Intelligence and Statistics},
  pages={1273--1282},
  year={2017},
  organization={PMLR}
}

@inproceedings{farhadkhani2022byzantine,
  title={Byzantine machine learning made easy by resilient averaging of momentums},
  author={Farhadkhani, Sadegh and Guerraoui, Rachid and Gupta, Nirupam and Pinot, Rafael and Stephan, John},
  booktitle={International Conference on Machine Learning},
  pages={6246--6283},
  year={2022},
  organization={PMLR}
}

@article{blanchard2017machine,
  title={Machine learning with adversaries: Byzantine tolerant gradient descent},
  author={Blanchard, Peva and El Mhamdi, El Mahdi and Guerraoui, Rachid and Stainer, Julien},
  journal={Advances in Neural Information Processing Systems},
  volume={30},
  year={2017}
}

@inproceedings{karimireddy2021learning,
  title={Learning from history for {B}yzantine robust optimization},
  author={Karimireddy, Sai Praneeth and He, Lie and Jaggi, Martin},
  booktitle={International Conference on Machine Learning},
  pages={5311--5319},
  year={2021},
  organization={PMLR}
}

@inproceedings{guerraoui2018hidden,
  title={The hidden vulnerability of distributed learning in \textsc{B}yzantium},
  author={Mhamdi, El Mahdi El and Guerraoui, Rachid and Rouault, S{\'e}bastien},
  booktitle={International Conference on Machine Learning},
  pages={3521--3530},
  year={2018},
  organization={PMLR}
}

@inproceedings{yin2018byzantine,
  title={Byzantine-robust distributed learning: Towards optimal statistical rates},
  author={Yin, Dong and Chen, Yudong and Kannan, Ramchandran and Bartlett, Peter},
  booktitle={International Conference on Machine Learning},
  pages={5650--5659},
  year={2018},
  organization={Pmlr}
}

@article{pillutla2022robust,
  title={Robust aggregation for federated learning},
  author={Pillutla, Krishna and Kakade, Sham M and Harchaoui, Zaid},
  journal={IEEE Transactions on Signal Processing},
  volume={70},
  pages={1142--1154},
  year={2022},
  publisher={IEEE}
}

@inproceedings{
karimireddy2022byzantinerobust,
title={Byzantine-Robust Learning on Heterogeneous Datasets via Bucketing},
author={Sai Praneeth Karimireddy and Lie He and Martin Jaggi},
booktitle={International Conference on Learning Representations},
year={2022}
}

@inproceedings{allouah2023fixing,
  title={Fixing by mixing: A recipe for optimal {B}yzantine {ML} under heterogeneity},
  author={Allouah, Youssef and Farhadkhani, Sadegh and Guerraoui, Rachid and Gupta, Nirupam and Pinot, Rafa{\"e}l and Stephan, John},
  booktitle={International Conference on Artificial Intelligence and Statistics},
  pages={1232--1300},
  year={2023},
  organization={PMLR}
}

@article{guerraoui2024byzantine,
  title={Byzantine machine learning: A primer},
  author={Guerraoui, Rachid and Gupta, Nirupam and Pinot, Rafael},
  journal={ACM Computing Surveys},
  volume={56},
  number={7},
  pages={1--39},
  year={2024},
  publisher={ACM New York, NY}
}

@book{guerraoui2024robust,
  title={Robust Machine Learning},
  author={Guerraoui, Rachid and Gupta, Nirupam and Pinot, Rafael},
  year={2024},
  publisher={Springer}
}

@book{tyler2008robust,
  title={Robust Statistics: Theory and Methods},
  author={Tyler, David E},
  year={2008},
  publisher={Taylor \& Francis}
}

@article{wang2022robust,
  title={Robust matrix completion based on factorization and truncated-quadratic loss function},
  author={Wang, Zhi-Yong and Li, Xiao Peng and So, Hing Cheung},
  journal={IEEE Transactions on Circuits and Systems for Video Technology},
  volume={33},
  number={4},
  pages={1521--1534},
  year={2023},
  publisher={IEEE}
}

@book{zoubir2018robust,
  title={Robust Statistics for Signal Processing},
  author={Zoubir, Abdelhak M and Koivunen, Visa and Ollila, Esa and Muma, Michael},
  year={2018},
  publisher={Cambridge University Press}
}

@article{wu2020federated,
  title={Federated variance-reduced stochastic gradient descent with robustness to {B}yzantine attacks},
  author={Wu, Zhaoxian and Ling, Qing and Chen, Tianyi and Giannakis, Georgios B},
  journal={IEEE Transactions on Signal Processing},
  volume={68},
  pages={4583--4596},
  year={2020},
  publisher={IEEE}
}

@article{baruch2019little,
  title={A little is enough: Circumventing defenses for distributed learning},
  author={Baruch, Gilad and Baruch, Moran and Goldberg, Yoav},
  journal={Advances in Neural Information Processing Systems},
  volume={32},
  year={2019}
}

@inproceedings{xie2020fall,
  title={Fall of empires: Breaking {B}yzantine-tolerant \textsc{SGD} by inner product manipulation},
  author={Xie, Cong and Koyejo, Oluwasanmi and Gupta, Indranil},
  booktitle={Uncertainty in Artificial Intelligence},
  pages={261--270},
  year={2020},
  organization={PMLR}
}

@article{dong2023byzantine,
  title={Byzantine-robust distributed online learning: Taming adversarial participants in an adversarial environment},
  author={Dong, Xingrong and Wu, Zhaoxian and Ling, Qing and Tian, Zhi},
  journal={IEEE Transactions on Signal Processing},
  pages={235-248},
  volume={72},
  year={2024},
  publisher={IEEE}
}

@inproceedings{yi2024near,
  title={Near-Optimal Resilient Aggregation Rules for Distributed Learning Using 1-Center and 1-Mean Clustering with Outliers},
  author={Yi, Yuhao and You, Ronghui and Liu, Hong and Liu, Changxin and Wang, Yuan and Lv, Jiancheng},
  booktitle={Proceedings of the AAAI Conference on Artificial Intelligence},
  volume={38},
  number={15},
  pages={16469--16477},
  year={2024}
}

@article{warnat2021swarm,
  title={Swarm learning for decentralized and confidential clinical machine learning},
  author={Warnat-Herresthal, Stefanie and Schultze, Hartmut and Shastry, Krishnaprasad Lingadahalli and Manamohan, Sathyanarayanan and Mukherjee, Saikat and Garg, Vishesh and Sarveswara, Ravi and H{\"a}ndler, Kristian and Pickkers, Peter and Aziz, N Ahmad and others},
  journal={Nature},
  volume={594},
  number={7862},
  pages={265--270},
  year={2021},
  publisher={Nature Publishing Group UK London}
}

@inproceedings{xie2019zeno,
  title={Zeno: Distributed stochastic gradient descent with suspicion-based fault-tolerance},
  author={Xie, Cong and Koyejo, Sanmi and Gupta, Indranil},
  booktitle={International Conference on Machine Learning},
  pages={6893--6901},
  year={2019},
  organization={PMLR}
}

@inproceedings{
allen2020byzantine,
title={Byzantine-Resilient Non-Convex Stochastic Gradient Descent},
author={Zeyuan Allen-Zhu and Faeze Ebrahimianghazani and Jerry Li and Dan Alistarh},
booktitle={International Conference on Learning Representations},
year={2021}
}

@article{rajput2019detox,
  title={\textsc{DETOX}: A redundancy-based framework for faster and more robust gradient aggregation},
  author={Rajput, Shashank and Wang, Hongyi and Charles, Zachary and Papailiopoulos, Dimitris},
  journal={Advances in Neural Information Processing Systems},
  volume={32},
  year={2019}
}

@article{data2020data,
  title={Data encoding for {B}yzantine-resilient distributed optimization},
  author={Data, Deepesh and Song, Linqi and Diggavi, Suhas N},
  journal={IEEE Transactions on Information Theory},
  volume={67},
  number={2},
  pages={1117--1140},
  year={2020},
  publisher={IEEE}
}

@inproceedings{li2019rsa,
  title={{RSA}: Byzantine-robust stochastic aggregation methods for distributed learning from heterogeneous datasets},
  author={Li, Liping and Xu, Wei and Chen, Tianyi and Giannakis, Georgios B and Ling, Qing},
  booktitle={Proceedings of the AAAI Conference on Artificial Intelligence},
  volume={33},
  number={01},
  pages={1544--1551},
  year={2019}
}

@article{JMLR:v26:24-1307,
  author  = {Jie Peng and Weiyu Li and Stefan Vlaski and Qing Ling},
  title   = {Mean Aggregator is More Robust than Robust Aggregators under Label Poisoning Attacks on Distributed Heterogeneous Data},
  journal = {Journal of Machine Learning Research},
  year    = {2025},
  volume  = {26},
  number  = {27},
  pages   = {1--51}
}

@inproceedings{zhao2024huber,
  title={A {H}uber loss minimization approach to {B}yzantine robust federated learning},
  author={Zhao, Puning and Yu, Fei and Wan, Zhiguo},
  booktitle={Proceedings of the AAAI Conference on Artificial Intelligence},
  volume={38},
  number={19},
  pages={21806--21814},
  year={2024}
}

@inproceedings{
dahan2024fault,
title={Fault Tolerant {ML}: Efficient Meta-Aggregation and Synchronous Training},
author={Tehila Dahan and Kfir Yehuda Levy},
booktitle={Forty-first International Conference on Machine Learning},
year={2024}
}

@inproceedings{yang2024byzantine,
  title={Byzantine-robust decentralized learning via remove-then-clip aggregation},
  author={Yang, Caiyi and Ghaderi, Javad},
  booktitle={Proceedings of the AAAI Conference on Artificial Intelligence},
  volume={38},
  number={19},
  pages={21735--21743},
  year={2024}
}

@article{luan2024robust,
  title={Robust federated learning: Maximum correntropy aggregation against \textsc{B}yzantine attacks},
  author={Luan, Zhirong and Li, Wenrui and Liu, Meiqin and Chen, Badong},
  journal={IEEE Transactions on Neural Networks and Learning Systems},
  volume={36},
  number={1},
  pages={62--75},
  year={2024},
  publisher={IEEE}
}

@inproceedings{zhang2025rethinking,
  title={Rethinking \textsc{B}yzantine Robustness in Federated Recommendation from Sparse Aggregation Perspective},
  author={Zhang, Zhongjian and Zhang, Mengmei and Wang, Xiao and Lyu, Lingjuan and Yan, Bo and Du, Junping and Shi, Chuan},
  booktitle={Proceedings of the AAAI Conference on Artificial Intelligence},
  volume={39},
  number={12},
  pages={13331--13338},
  year={2025}
}

@inproceedings{wu2024fedbiot,
  title={{FedBiOT}: \textsc{LLM} local fine-tuning in federated learning without full model},
  author={Wu, Feijie and Li, Zitao and Li, Yaliang and Ding, Bolin and Gao, Jing},
  booktitle={Proceedings of the 30th ACM SIGKDD Conference on Knowledge Discovery and Data Mining},
  pages={3345--3355},
  year={2024}
}

@article{huynh2025certified,
  title={Certified unlearning for federated recommendation},
  author={Huynh, Thanh Trung and Nguyen, Trong Bang and Nguyen, Thanh Toan and Nguyen, Phi Le and Yin, Hongzhi and Nguyen, Quoc Viet Hung and Nguyen, Thanh Tam},
  journal={ACM Transactions on Information Systems},
  volume={43},
  number={2},
  pages={1--29},
  year={2025},
  publisher={ACM New York, NY}
}

@inproceedings{fang2020local,
  title={Local model poisoning attacks to \textsc{B}yzantine-robust federated learning},
  author={Fang, Minghong and Cao, Xiaoyu and Jia, Jinyuan and Gong, Neil},
  booktitle={29th USENIX security symposium (USENIX Security 20)},
  pages={1605--1622},
  year={2020}
}

@article{so2020byzantine,
  title={Byzantine-resilient secure federated learning},
  author={So, Jinhyun and G{\"u}ler, Ba{\c{s}}ak and Avestimehr, A Salman},
  journal={IEEE Journal on Selected Areas in Communications},
  volume={39},
  number={7},
  pages={2168--2181},
  year={2021},
  publisher={IEEE}
}

@inproceedings{bao2024boba,
  title={\textsc{BOBA}: Byzantine-robust federated learning with label skewness},
  author={Bao, Wenxuan and Wu, Jun and He, Jingrui},
  booktitle={International Conference on Artificial Intelligence and Statistics},
  pages={892--900},
  year={2024},
  organization={PMLR}
}

@inproceedings{fedin2023byzantine,
  title={Byzantine-robust loopless stochastic variance-reduced gradient},
  author={Fedin, Nikita and Gorbunov, Eduard},
  booktitle={International Conference on Mathematical Optimization Theory and Operations Research},
  pages={39--53},
  year={2023},
  organization={Springer}
}

@inproceedings{zhu2022byzantine,
  title={Byzantine-robust aggregation with gradient difference compression and stochastic variance reduction for federated learning},
  author={Zhu, Heng and Ling, Qing},
  booktitle={ICASSP 2022-2022 IEEE International Conference on Acoustics, Speech and Signal Processing (ICASSP)},
  pages={4278--4282},
  year={2022}
}

@inproceedings{he2016deep,
  title={Deep residual learning for image recognition},
  author={He, Kaiming and Zhang, Xiangyu and Ren, Shaoqing and Sun, Jian},
  booktitle={Proceedings of the IEEE Conference on Computer Vision and Pattern Recognition},
  pages={770--778},
  year={2016}
}

@misc{lecun2010mnist,
  title={{MNIST} handwritten digit database},
  author={LeCun, Yann and Cortes, Corinna and Burges, Chris and others},
  year={2010},
  publisher={Florham Park, NJ, USA}
}

@article{krizhevsky2009learning,
  title={Learning multiple layers of features from tiny images},
  author={Krizhevsky, Alex and Hinton, Geoffrey and others},
  year={2009},
  publisher={Toronto, ON, Canada}
}

@article{xiao2017fashion,
  title={Fashion-{MNIST}: a novel image dataset for benchmarking machine learning algorithms},
  author={Xiao, Han and Rasul, Kashif and Vollgraf, Roland},
  journal={arXiv preprint arXiv:1708.07747},
  year={2017}
}

@article{he2025ppbr,
  title={\textsc{PPBR}: Privacy-Preserving and \textsc{B}yzantine-robust Edge-Assisted Hierarchical Federated Learning in Mobile Networks},
  author={He, Yuanyuan and Zhang, Jie and Yang, Peng and Sun, Zhe and Shen, Xuemin},
  journal={IEEE Transactions on Mobile Computing},
  year={2025},
  publisher={IEEE}
}

@inproceedings{huang2024self,
  title={Self-driven entropy aggregation for \textsc{B}yzantine-robust heterogeneous federated learning},
  author={Huang, Wenke and Shi, Zekun and Ye, Mang and Li, He and Du, Bo},
  booktitle={Forty-first International Conference on Machine Learning},
  year={2024}
}

@article{zuo2025federated,
  title={Federated Learning Resilient to \textsc{B}yzantine Attacks and Data Heterogeneity},
  author={Zuo, Shiyuan and Yan, Xingrun and Fan, Rongfei and Hu, Han and Shan, Hangguan and Quek, Tony QS and Zhao, Puning},
  journal={IEEE Transactions on Mobile Computing},
  year={2025},
  publisher={IEEE}
}

@article{chen2023energy,
  title={Energy efficient and differentially private federated learning via a piggyback approach},
  author={Chen, Rui and Huang, Chenpei and Qin, Xiaoqi and Ma, Nan and Pan, Miao and Shen, Xuemin},
  journal={IEEE Transactions on Mobile Computing},
  volume={23},
  number={4},
  pages={2698--2711},
  year={2023},
  publisher={IEEE}
}

@article{huang2024pprp,
  title={{PPRP}: preserving location privacy for range-based positioning in mobile networks},
  author={Huang, Cheng and Liu, Dongxiao and Yang, Anjia and Lu, Rongxing and Shen, Xuemin},
  journal={IEEE Transactions on Mobile Computing},
  volume={23},
  number={10},
  pages={9451--9468},
  year={2024},
  publisher={IEEE}
}

@ARTICLE{11241044,
  author={Wang, Zhi-Yong and Nan Sheng, Hao and Yang, Qiushi and Cheung So, Hing},
  journal={IEEE Transactions on Cybernetics}, 
  title={Order-Optimal Byzantine-Robust Learning Under Heterogeneity via Fair Gradient Clipping}, 
  year={2026},
  volume={56},
  number={3},
  pages={1503-1514},
  doi={10.1109/TCYB.2025.3628486}}

@ARTICLE{11303260,
  author={Wang, Zhi-Yong and Sheng, Hao Nan and So, Hing Cheung and Sun, Jiande and Song, Linqi and Xu, Weitao},
  journal={IEEE Transactions on Circuits and Systems for Video Technology}, 
  title={Robust Federated Learning under Heterogeneity via Rank-One and Column-Sparsity Model}, 
  year={2025},
  volume={},
  number={},
  pages={1-1},
  doi={10.1109/TCSVT.2025.3645323}}

\clearpage

\end{document}